\documentclass[11pt]{article}

\usepackage[final]{acl}

\usepackage{times}
\usepackage{latexsym}

\usepackage[T1]{fontenc}
\usepackage[utf8]{inputenc}

\usepackage{microtype}

\usepackage{inconsolata}

\usepackage{graphicx}
\usepackage{amsmath}
\usepackage{amsfonts}
\usepackage{algorithm}
\usepackage{algorithmic}
\usepackage{enumitem}
\usepackage{multirow}
\usepackage{colortbl}
\usepackage{xcolor}
\usepackage{booktabs}
\usepackage{hyperref}
\usepackage{tcolorbox}
\usepackage{tabularx}
\usepackage[normalem]{ulem}
\usepackage{soul}
\usepackage{makecell}
\usepackage{listings}

\title{One Refiner to Unlock Them All: \\Inference-Time Reasoning Elicitation via Reinforcement Query Refinement}
\author{
 \textbf{Yixiao Zhou\textsuperscript{1,2}},
 \textbf{Dongzhou Cheng\textsuperscript{2,3}},
 \textbf{Zhiliang Wu\textsuperscript{1}},
\\
 \textbf{Yi Yang\textsuperscript{1}},
 \textbf{Yu Cheng\textsuperscript{2,4}},
 \textbf{Hehe Fan\textsuperscript{1}}
 \\
 \\
 \textsuperscript{1}Zhejiang University,
 \textsuperscript{2}Shanghai Innovation Institute,
\\
 \textsuperscript{3}Southeast University,
 \textsuperscript{4}The Chinese University of Hong Kong
 \\
 \texttt{\{12421181,wu\_zhiliang,yangyics,hehefan\}@zju.edu.cn},
\\
 \texttt{230249457@seu.edu.cn},
 \texttt{chengyu@cse.cuhk.edu.hk}
\\
 \small{
   \textbf{Correspondence:} \href{mailto:chengyu@cse.cuhk.edu.hk}{chengyu@cse.cuhk.edu.hk}, \href{mailto:hehefan@zju.edu.cn}{hehefan@zju.edu.cn}
 }
}

\begin{document}
\maketitle
\begin{abstract}
Large Language Models (LLMs) often fail to utilize their latent reasoning capabilities due to a distributional mismatch between ambiguous human inquiries and the structured logic required for machine activation. Existing alignment methods either incur prohibitive $O(N)$ costs by fine-tuning each model individually or rely on static prompts that fail to resolve query-level structural complexity. In this paper, we propose ReQueR (\textbf{Re}inforcement \textbf{Que}ry \textbf{R}efinement), a modular framework that treats reasoning elicitation as an inference-time alignment task. We train a specialized Refiner policy via Reinforcement Learning to rewrite raw queries into explicit logical decompositions, treating frozen LLMs as the environment. Rooted in the classical Zone of Proximal Development from educational psychology, we introduce the Adaptive Solver Hierarchy, a curriculum mechanism that stabilizes training by dynamically aligning environmental difficulty with the Refiner's evolving competence. ReQueR yields consistent absolute gains of 1.7\%--7.2\% across diverse architectures and benchmarks, outperforming strong baselines by 2.1\% on average. Crucially, it provides a promising paradigm for one-to-many inference-time reasoning elicitation, enabling a single Refiner trained on a small set of models to effectively unlock reasoning in diverse unseen models. Code is available at \url{https://github.com/newera-xiao/ReQueR}.
\end{abstract}

\section{Introduction}
\begin{figure}[t]
    \centering
    \includegraphics[width=1\linewidth]{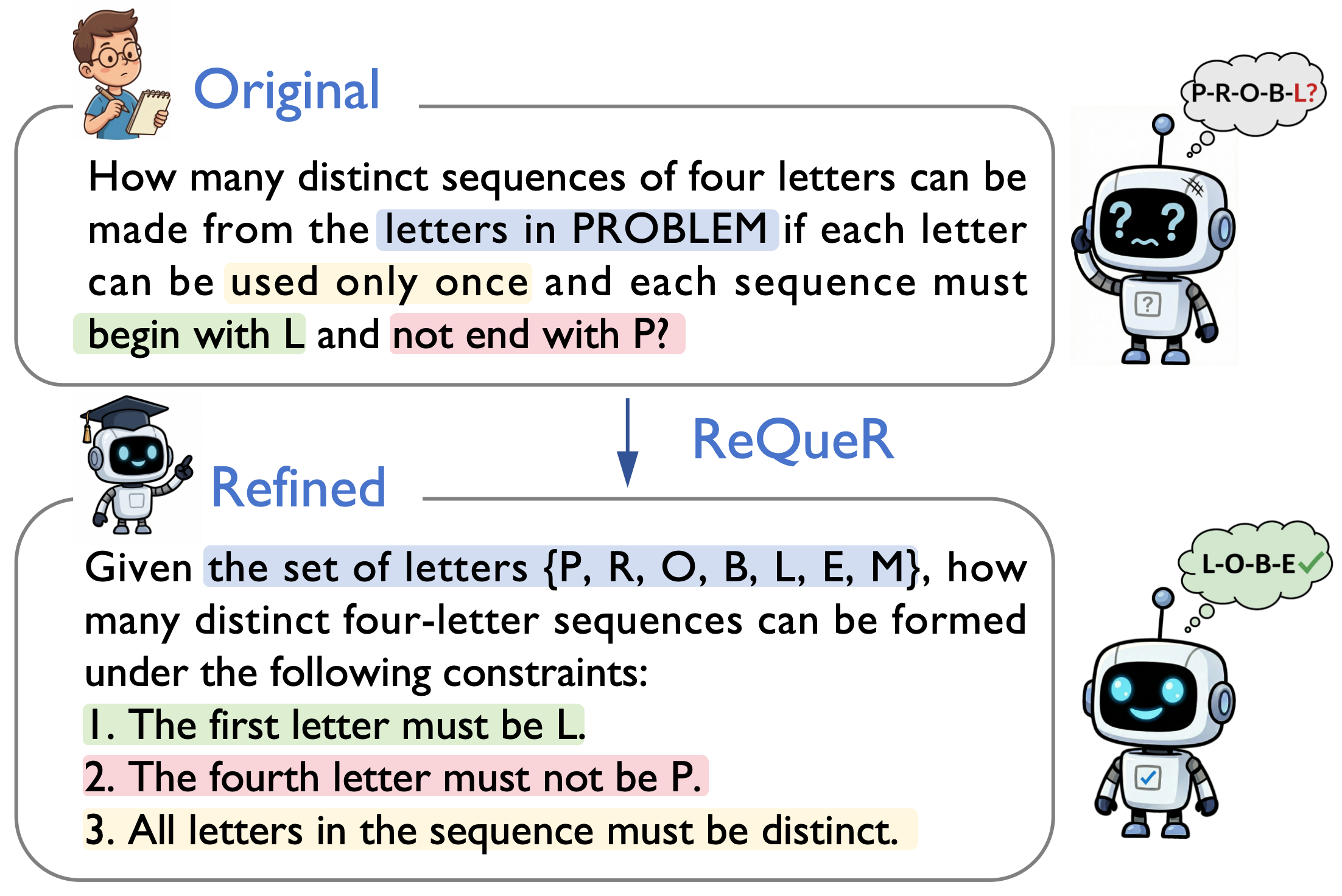}
    \caption{\textbf{Eliciting latent reasoning via inference-time query refinement.} A Solver struggles with the ambiguity of a raw human query (top). ReQueR functions as an optimization frontend, reformulating the input into a structured, explicit format (bottom). By resolving ambiguity and employing strategies like variable-mapping, the Refiner aligns the query with the Solver's reasoning patterns, effectively unlocking its latent capabilities without requiring parameter updates.}
    \label{fig:intro_demo}
    \vspace{-5pt}
\end{figure}

% [Para 1: The Context & Gap]
Large Language Models (LLMs) possess significant reasoning potential~\cite{shao2024deepseekmath}, yet a gap remains between their intrinsic knowledge and actual performance. Their reasoning capabilities are often not fully activated and are highly sensitive to input phrasing, where small perturbations can lead to failure~\cite{zhao2021calibrate,mirzadeh2024gsm,cai2025does,an2025amo}. We argue that this bottleneck stems from a distributional mismatch: while human inquiries are often ambiguous and noisy, LLM reasoning engines require explicit and structured logic to function optimally~\cite{deng2023rephrase}. Therefore, the key to unlocking latent reasoning lies in bridging this alignment gap between human intent and machine activation.

Bridging this gap, however, is far from trivial.
% First, parameter-update methods, such as Supervised Fine-Tuning (SFT)~\cite{hu2022lora} and Reinforcement Learning from Verifiable Rewards (RLVR)~\cite{guo2025deepseek}, attempt to align models by modifying internal weights. While effective, these approaches suffer from poor scalability and prohibitive costs: aligning $N$ distinct models necessitates $N$ independent training runs, leading to a high $O(N)$ complexity. Furthermore, they are inapplicable to proprietary black-box models where gradient access is restricted.
First, parameter-update methods, such as Supervised Fine-Tuning (SFT)~\cite{hu2022lora,xiaometa,chang2026balora,dong2025aurora} and Reinforcement Learning from Verifiable Rewards (RLVR)~\cite{guo2025deepseek}, attempt to align models by modifying internal weights. While effective, these approaches are impractical for agent systems where models are frequently swapped or upgraded. Aligning $N$ models necessitates $N$ independent training runs, leading to a $O(N)$ complexity. Furthermore, they are inapplicable to proprietary black-box models where gradient access is restricted.
Alternatively, while prompt engineering offers a lightweight substitute, most existing strategies act as static wrappers that merely append shallow instructions to the input~\cite{wei2022chain}. They often fail to perform the deep rewriting required to simplify complex query structures. Moreover, automated prompt optimization tends to be overfitted to specific models or datasets, severely limiting their generalization to unseen environments~\cite{yang2023large}.
Consequently, existing paradigms fail to provide a universal solution at inference time, as the optimal reasoning trigger remains highly task-dependent and varies significantly across divergent model architectures.

% [Para 3: The Proposal - ReQueR]
To address these gaps, we introduce ReQueR (\textbf{Re}inforcement \textbf{Que}ry \textbf{R}efinement), an inference-time frontend that reformulates human queries to activate the reasoning capabilities of downstream LLMs. Instead of per-model fine-tuning or task-level prompt optimization, we use reinforcement learning to train a Refiner LLM as a unified meta-policy. As illustrated in Figure~\ref{fig:intro_demo}, ReQueR performs query-level reformulation, translating ambiguous human intent into structured machine logic.
During training, the Refiner is optimized through reward signals from frozen Solver LLMs responding to the reformulated queries. This approach shifts the alignment burden from the $O(N)$ cost of independent model-tuning to a scalable $O(1)$ inference-time paradigm, while offering a granularity that static prompt optimization lacks. Once trained, a single Refiner can unlock the reasoning potential of diverse, unseen models without any additional parameter updates. Unlike dataset-level prompt optimization, ReQueR reformulates each query individually, ensuring that the reasoning trigger is tailored to the specific logic of every sample.

% [Para 4: The Challenge]
However, training this Refiner policy presents significant challenges. The main problem is reward sparsity~\cite{yu2025dapo,ding2026prpo,li2025choice,yu2026unveiling,li2026dyjr} caused by the use of static training environments. Solvers that are either too weak or overly capable fail to provide informative feedback, often driving the Refiner to exploit model-specific biases rather than learning general reasoning logic. Another problem is reward hacking, where the Refiner learns to cheat by directly leaking the answer into the refined query~\cite{gao2023scaling}. While this results in high training scores, it leads to catastrophic failure when the Refiner is applied to new, unseen environments.

% [Para 5: The Innovation (ASH & PPL Judge)]
To address these challenges, we introduce two mechanisms. First, the Adaptive Solver Hierarchy (ASH) mitigates reward sparsity and model-specific overfitting. Grounded in the Zone of Proximal Development~\cite{vygotsky1978mind}, ASH utilizes a pool of Solvers of varied capabilities and performs per-sample dynamic matching according to Refiner's current competence. By escalating Solver difficulty upon consistent failure and de-escalating when success becomes trivial, ASH ensures the Refiner evolves within its optimal growth zone, maintaining a dense reward signal~\cite{bengio2009curriculum,xiao2026points,li2025curriculum} without model-specific dependencies. Second, we implement a perplexity-based constraint to suppress reward hacking. Unlike computationally expensive and unstable generative LLM judges~\cite{wang2024large}, our approach uses a single forward pass to detect potential leakage, enabling quick monitoring within the RL loop. 

% [Para 6: Contributions]
Our contributions can be summarized as follows:
\begin{itemize}
    \item We propose \textbf{ReQueR}, a modular inference-time Query Refinement framework. By learning a unified meta-policy for per-sample query refinement, ReQueR unlock the reasoning potential of diverse LLMs.
    \item We introduce the Adaptive Solver Hierarchy alongside an efficient leakage detection constraint. ASH addresses reward sparsity through per-sample dynamic Solver selection tailored to the Refiner’s refined queries, while our perplexity-based constraint effectively mitigates reward hacking.
    \item Extensive experiments across various LLMs and reasoning tasks demonstrate that ReQueR consistently improves reasoning performance, and outperforms existing baselines. Notably, we demonstrate one-to-many generalization, where a single Refiner trained on small-scale open-source models successfully unlocks the latent reasoning potential of unseen models.
\end{itemize}

% [Para 2: The Critique]
\begin{figure*}[t]
    \centering
    \includegraphics[width=0.85\textwidth]{RL_Arch.png} 
    \caption{\textbf{ReQueR reinforcement learning pipeline.} The Refiner generates multiple refined queries for a Solver dynamically assigned by the ASH based on problem difficulty. After computing rewards from the Solver's answers, the framework optimizes the Refiner policy and updates Solver labels to ensure a stable learning curriculum.}
    \label{fig:rl_arch}
    \vspace{-10pt}
\end{figure*}

\section{Related Work}
\paragraph{Input Optimization for LLMs} LLM performance is highly sensitive to prompt quality, driving a shift from manual engineering to automated optimization. Frameworks like TextGrad~\cite{yuksekgonul2024textgrad} and OPRO~\cite{yang2023large} utilize gradient-like feedback to refine prompts, while GEPA~\cite{agrawal2025gepa}  employs genetic evolution. However, these methods are often task- and model-dependent, lacking the flexibility for per-sample dynamic adaptation . Alternatively, rewriting strategies such as RaR~\cite{deng2023rephrase} and RE2~\cite{xu2024re} rely on static, heuristic-based templates. ReQueR employs Reinforcement Learning to learn a universal meta-policy that reconstructs a problem's logical skeleton. As an adaptive inference-time frontend, it enables $O(1)$ reasoning elicitation, unlocking the latent potential of unseen Solvers and tasks.

\paragraph{Reinforcement Learning for Reasoning} Reinforcement Learning (RL) has emerged as a dominant paradigm for enhancing the reasoning capabilities of LLMs~\cite{gulcehre2023reinforced}. Reinforcement Learning with Verifiable Rewards (RLVR)~\cite{guo2025deepseek} leverages objective ground-truth feedback, such as mathematical correctness~\cite{cobbe2021training,lin2026cec} or code execution results~\cite{le2022coderl}, to provide precise supervision. Group Relative Policy Optimization~\cite{shao2024deepseekmath,yao2024swift} further improved training efficiency through group-relative advantage normalization, eliminating the need for a separate value-function critic. A broader discussion of recent RL-for-reasoning advances is provided in Appendix~\ref{appendix:extended_rl}. Unlike these methods that enhance Solvers directly, ReQueR optimizes a Refiner meta-policy for one-to-many reasoning elicitation, unlocking the latent potential of diverse frozen Solvers at inference time.

\section{The ReQueR Framework}

\subsection{Problem Formulation}
We reformulate the reasoning elicitation task as an \textbf{inference-time alignment} problem. Unlike SFT or RL that modifies the internal parameters of a model to match human preferences, ReQueR aligns the input distribution to match the intrinsic reasoning triggers of frozen large language models.

\paragraph{The Refiner and the Solver Set} 
Let $\mathcal{D} = \{(x, y^*)\}$ be a dataset of reasoning problems where $x$ represents the raw human inquiry and $y^*$ is the ground-truth answer. We define the \textbf{Refiner} as a policy $\pi_\theta(x' | x)$, parameterized by $\theta$, which transforms $x$ into a structured, refined query $x'$. The environment consists of a heterogeneous set of frozen \textbf{Solvers} $\mathcal{S} = \{\mathcal{M}_k\}_{k=1}^K$, ranked by their intrinsic reasoning capabilities. Each Solver $\mathcal{M}_k \in \mathcal{S}$ is treated as a black box that, given an input $q$, generates a reasoning trajectory $\tau$ and a predicted answer $\hat{y} = \text{Extract}(\mathcal{M}_k(q))$.

\paragraph{The $O(1)$ Alignment Objective} 
The core ambition of ReQueR is to learn a universal \textbf{meta-policy} that generalizes across various Solvers without requiring individual parameter updates. To formalize this ``one-to-many'' alignment, we define the Cross-Model Generalization performance for a given problem $(x, y^*)$ as:
\begin{equation}
\mathcal{G}(\theta; x, y^*) = \mathbb{E}_{\mathcal{M} \in \mathcal{S}} [ \mathcal{R}(y^*, \mathcal{M}(\pi_\theta(x))) ].
\label{eq:objective_function}
\end{equation}
The global objective is then formulated as:
\begin{equation}
\max_\theta \mathbb{E}_{(x, y^*) \sim \mathcal{D}} \left[ \mathcal{G}(\theta; x, y^*) \right],
\end{equation}
where $\mathcal{R}$ is the composite reward function (Section~\ref{sec:reward}), and the expectation over $\mathcal{S}$ is managed by the Adaptive Solver Hierarchy (Section~\ref{sec:ash}).

\paragraph{Decoupling and Scalability} ReQueR shifts the alignment burden from parameters to input optimization, replacing $O(N)$ model-specific tuning with a scalable $O(1)$ paradigm. By optimizing across a Solver manifold, the Refiner internalizes universal reasoning triggers that transcend individual architectures. Once trained, $\pi_\theta$ can be deployed to unlock latent potential in unseen Solvers ($\mathcal{M}_{\text{unseen}}$) at inference time with zero additional training cost (Theoretical analysis in Appendix~\ref{sec:theory}).

\paragraph{Comparison with Prompt Optimization} ReQueR transcends Automatic Prompt Optimization (APO) in two critical dimensions. First, while APO searches for a static, task-level wrapper string, ReQueR learns a generative meta-policy that produces a unique, tailored refinement for every instance. Second, by optimizing against a Solver manifold rather than a single model, ReQueR internalizes model-agnostic reasoning triggers. Detailed comparisons are provided in Appendix~\ref{appendix:paradigm_comparison}

\subsection{Environmental Dynamics via ASH}
\label{sec:ash}

To optimize the objective defined in Equation~\eqref{eq:objective_function}, the Refiner must interact with the Solver distribution $\mathcal{S}$. However, training against a fixed Solver often leads to \textit{reward sparsity}. A weak Solver may fail consistently regardless of query quality, while a dominant Solver might succeed even with suboptimal queries. To maintain a high-quality feedback loop, we propose the Adaptive Solver Hierarchy (ASH), a curriculum mechanism that dynamically manages environmental complexity.

\paragraph{Dynamic Difficulty Adjustment}
We rank the Solvers in $\mathcal{S}$ by capability, $\mathcal{S} = \{\mathcal{M}_1, \dots, \mathcal{M}_K\}$. Each sample $x_i \in \mathcal{D}$ is associated with a local difficulty index $k_i \in \{1, \dots, K\}$. During training, the Refiner generates a group of $G$ queries $\{x'_{i,j}\}_{j=1}^G$. Let $S_i = \sum_{j=1}^G \mathbb{I}[\hat{y}_{i,j} = y^*_i]$ denote the success count within the group. To maintain an informative reward signal, $k_i$ is updated for the next iteration:
\begin{equation}
k_i^{(t+1)} = 
\begin{cases} 
\label{eq:ash}
\min(K, k_i^{(t)} + 1) & \text{if } S_i = 0, \\
\max(1, k_i^{(t)} - 1) & \text{if } S_i = G, \\
k_i^{(t)} & \text{otherwise.}
\end{cases}
\end{equation}

\begin{algorithm}[t]
\caption{ReQueR RL Training Procedure}
\label{alg:requer}
\begin{algorithmic}[1]
\REQUIRE $\mathcal{D}$, $\pi_{\theta}$, $\pi_{\text{ref}}$, $\{\mathcal{M}_k\}_{1}^K$, $\mathcal{M}_{j}$
\ENSURE Optimized Policy $\pi_{\theta^*}$
\STATE \textbf{Init:} $\theta \leftarrow \theta_{\text{SFT}}$, $k_i \leftarrow \lceil K/2 \rceil \ \forall x_i \in \mathcal{D}$

\FOR{epoch $e = 1, \dots, E$}
    \STATE Sample mini-batch $\mathcal{B} \subseteq \mathcal{D}$
    \FOR{each $(x_i, y^*_i) \in \mathcal{B}$}
        \STATE \textcolor{gray}{\textit{// 1. Group Sampling \& Evaluation}}
        \STATE $\{x'_{i,j}\}_{j=1}^G \sim \pi_{\theta}(\cdot | x_i)$
        \STATE $\hat{y}_{i,j} \leftarrow \text{Extract}(\mathcal{M}_{k_i}(x'_{i,j})), \forall j$
        \STATE $R_{i,j} \leftarrow \mathbb{I}[\hat{y}_{i,j} = y^*_i] - \lambda R_{\text{leak}}(x'_{i,j}, x_i)$
        
        \STATE \textcolor{gray}{\textit{// 2. ASH Update via Equation~\eqref{eq:ash}}}
        \STATE Let $S_i = \sum_{j} \mathbb{I}[\hat{y}_{i,j} = y^*_i]$
        \STATE $\Delta k_i \leftarrow \mathbb{I}[S_i = 0] - \mathbb{I}[S_i = G]$
        \STATE $k_i \leftarrow \text{clamp}(k_i + \Delta k_i, 1, K)$
        
        \STATE \textcolor{gray}{\textit{// 3. Group Advantage Normalization}}
        \STATE $A_{i,j} \leftarrow (R_{i,j} - \bar{R}_i) / (\sigma(R_i) + \epsilon)$
    \ENDFOR
    \STATE \textcolor{gray}{\textit{// 4. Gradient Update via Equation~\eqref{eq:grpo}}}
    \STATE $\theta \leftarrow \theta + \eta \nabla_\theta \mathcal{J}(\theta; \mathcal{B})$
\ENDFOR
\end{algorithmic}
\end{algorithm}
\vspace{10pt}

\paragraph{ASH as Feedback Calibration}
This mechanism is conceptually grounded in the \textit{Zone of Proximal Development} (ZPD), which posits that optimal learning occurs when the environment's difficulty is precisely calibrated to the agent's current competence. ASH operationalizes this by addressing two extremes of gradient degradation~\cite{zhang2026logical}.
First, when $S_i = 0$, the environment is effectively ``deaf'' to the Refiner's structural improvements, leading to a total absence of positive reinforcement. In this scenario, ASH escalates to a more capable Solver to detect whether the refined query contains valid reasoning cues that only a stronger model can decode.
Conversely, when $S_i = G$, the task is trivialized by an overly powerful Solver that bypasses the need for high-quality input. Here, ASH then de-escalates to impose ``structural pressure'', forcing the Refiner to generate more explicit and robust logic.
By dynamically maintaining the Refiner within its ZPD, ASH prevents the policy from overfitting to a specific model's idiosyncratic tolerance and facilitates the discovery of universal reasoning triggers across the entire Solver manifold.

\subsection{Reward Design}
\label{sec:reward}

To ensure the Refiner learns to elicit reasoning rather than simply injecting answers, we define a composite reward $R = R_{\text{acc}} - \lambda \cdot R_{\text{leak}}$, which balances task performance with the integrity of the reasoning process.

\paragraph{Task Success ($R_{\text{acc}}$)}
The primary objective is the correctness of the Solver's predicted answer $\hat{y}$ compared to the ground truth $y^*$:
\begin{equation}
R_{\text{acc}} = \mathbb{I}[\hat{y} = y^*].
\end{equation}

\paragraph{Leak Penalty ($R_{\text{leak}}$)}
Information leakage occurs when the Refiner provides trivial shortcuts that bypass logical complexity. To detect such exploitation of non-reasoning shortcuts, we utilize a judge model $\mathcal{M}_{j}$ to measure the conditional perplexity $\text{PPL}(y^* | q)$ of the answer sequence $y^* = (t_1, \dots, t_L)$:
\begin{equation}
\text{PPL}(y^* | q) = \exp \left( -\frac{1}{L} \log P_{\mathcal{M}_{j}}(y^* | q) \right).
\end{equation}
We quantify leakage severity via the Perplexity Drop Ratio. To suppress reward hacking, we apply an indicator-based penalty $R_{\text{leak}}$ that triggers when the relative shift in the answer's predictability before and after refinement exceeds a tolerance threshold $\tau_{\text{leak}}$:
\begin{equation}
R_{\text{leak}} = \mathbb{I} \left[ \frac{\text{PPL}(y^* | x)}{\text{PPL}(y^* | x') + \epsilon} > \tau_{\text{leak}} \right].
\end{equation}
This constraint acts as a semantic guardrail, compelling the Refiner to prioritize structural disambiguation over leaking surface-level cues, thereby ensuring the elicitation of genuine latent reasoning.

\subsection{Training Pipeline}

The training of ReQueR proceeds in two distinct stages: a supervised warm-up to establish a baseline alignment capability, followed by reinforcement learning to refine the meta-policy (Figure~\ref{fig:rl_arch}).

\paragraph{Phase 1: Cold-Start Bootstrapping}
To stabilize subsequent reinforcement learning, we initialize the Refiner $\pi_\theta$ via Supervised Fine-Tuning on a curated dataset $\mathcal{D}_{\text{SFT}}$ containing 6,144 high-quality $(x, x')$ pairs. These samples are synthesized by \text{DeepSeek-R1}~\cite{guo2025deepseek} and further refined through rigorous manual curation to ensure logical precision. The primary objective of this phase is to align the Refiner with the required output format and establish preliminary query-refining capabilities. The policy is warmed up by minimizing the standard negative log-likelihood:
\begin{equation}
\mathcal{L}_{\text{SFT}}(\theta) = - \mathbb{E}_{(x, x') \sim \mathcal{D}_{\text{SFT}}} [\log \pi_\theta(x' | x)].
\end{equation}

% \paragraph{Phase 2: Online Optimization}
% Building upon the initialized policy, we employ Group Relative Policy Optimization\cite{guo2025deepseek} to evolve the Refiner. The advantage $A_j$ for each refinement is computed via group normalization of the rewards:
% \begin{equation}
% A_j = \frac{R^{(j)} - \text{mean}(\{R^{(i)}\}_{i=1}^G)}{\text{std}(\{R^{(i)}\}_{i=1}^G) + \epsilon}.
% \end{equation}
% The Refiner is then optimized by maximizing a clipped surrogate objective with a KL penalty. The global objective is formulated as:
% \begin{equation}
% \mathcal{J}(\theta) = \mathbb{E}_{x \sim \mathcal{D}} \bigg[ \frac{1}{G} \sum_{j=1}^G \mathcal{L}_j^{\text{clip}} - \beta \mathbb{D}_{KL} \bigg],
% \end{equation}
% where $\mathcal{L}_j^{\text{clip}}$ is the clipped surrogate loss:
% \begin{equation}
% \mathcal{L}_j^{\text{clip}} = \min \left( r_j A_j, \text{clip}(r_j, 1 \pm \epsilon) A_j \right).
% \end{equation}

% Here, $r_j = \pi_\theta(x'_j|x) / \pi_{\theta_{old}}(x'_j|x)$ is the probability ratio, and the $\text{clip}$ function restricts $r_j$ to the interval $[1-\epsilon, 1+\epsilon]$. This process enables the Refiner to master universal reasoning triggers across the ASH hierarchy.

\paragraph{Phase 2: Online Optimization}
Building upon the initialized policy, we employ Group Relative Policy Optimization~\cite{guo2025deepseek} to evolve the Refiner. The advantage $A_j$ for each refinement is computed via group normalization of the rewards:
\begin{equation}
A_j = \frac{R^{(j)} - \text{mean}(\{R^{(i)}\}_{i=1}^G)}{\text{std}(\{R^{(i)}\}_{i=1}^G) + \epsilon}.
\end{equation}
The Refiner is then optimized by maximizing a clipped surrogate objective with a KL penalty. The global objective is formulated as:
\begin{equation}
\begin{aligned}
\label{eq:grpo}
& \mathcal{J}(\theta) = \mathbb{E}_{x \sim \mathcal{D}} \bigg[ \frac{1}{G} \sum_{j=1}^G \min \Big( r_j A_j, \text{clip}(r_j, \\
& \quad 1-\epsilon, 1+\epsilon) A_j \Big) - \beta \mathbb{D}_{KL}(\pi_\theta \| \pi_{\text{ref}}) \bigg],
\end{aligned}
\end{equation}
where $r_j = \pi_\theta(x'_j|x) / \pi_{\theta_{\text{old}}}(x'_j|x)$ is the probability ratio, and the $\text{clip}$ function restricts $r_j$ to the interval $[1-\epsilon, 1+\epsilon]$. This process enables the Refiner to explore universal reasoning triggers. The complete procedure is summarized in Algorithm~\ref{alg:requer}.

\definecolor{darkgreen}{rgb}{0.0, 0.5, 0.0}
\newcommand{\gain}[1]{{\,\textcolor{darkgreen}{\footnotesize{(+#1)}}}}

\begin{table*}[t]
    \centering
    \caption{\textbf{Performance comparison on extended reasoning benchmarks.} ReQueR (Qwen3-4B) is trained on a fixed ASH (Qwen3-0.6B/1.7B/4B) and directly applied as a plug-and-play Refiner to all other models. This evaluates the Refiner's ability to elicit reasoning from unseen architectures. \textbf{Bold} indicates best performance; \textcolor{darkgreen}{green} values denote gains over zero-shot CoT.}
    \vspace{-5pt}
    \label{tab:main_results1}
    
    \small 
    \setlength{\tabcolsep}{5pt} % 垂直换行后，可以适当调大列间距，让表格更疏朗
    
    \begin{tabular}{llcccccccc}
        \toprule
        \textbf{Model} & \textbf{Method} & \makecell[c]{\textbf{AMC23}} & \makecell[c]{\textbf{Olym.}\\\textbf{Bench}} & \makecell[c]{\textbf{Omni-}\\\textbf{MATH}} & \makecell[c]{\textbf{GSM-}\\\textbf{Plus}} & \makecell[c]{\textbf{GSM-}\\\textbf{Symb.}} & \makecell[c]{\textbf{MATH-}\\\textbf{500}} & \makecell[c]{\textbf{GPQA-}\\\textbf{Diam.}} & \makecell[c]{\textbf{Avg.}} \\
        \midrule
        
        % --- Llama Models ---
        \multicolumn{10}{c}{\textit{\textbf{Llama Models}}} \\
        \cmidrule[0.25pt](lr){1-10}
        
        \multirow{2}{*}{Llama-3.2-3B-Instruct} 
        & CoT & 19.69 & 12.76 & 11.04 & 55.43 & 60.64 & 46.60 & 26.77 & 33.28 \\
        & \textbf{ReQueR} & \textbf{22.19} & \textbf{19.88} & \textbf{13.48} & \textbf{65.76} & \textbf{69.38} & \textbf{50.20} & \textbf{30.30} & \textbf{38.74}\gain{5.47} \\ 
        \cmidrule[0.25pt](lr){1-10}

        \multirow{2}{*}{Llama-3.1-8B-Instruct} 
        & CoT & 25.94 & 16.02 & 11.63 & 66.86 & 74.44 & 45.80 & 27.78 & 38.35 \\
        & \textbf{ReQueR} & \textbf{30.94} & \textbf{21.81} & \textbf{14.02} & \textbf{71.76} & \textbf{75.82} & \textbf{54.00} & \textbf{31.31} & \textbf{42.81}\gain{4.46} \\ 
        \cmidrule[0.25pt](lr){1-10}

        \multirow{2}{*}{Llama-3.1-70B-Instruct} 
        & CoT & 50.31 & 31.31 & 17.93 & 80.81 & 85.76 & 64.60 & 47.73 & 54.06 \\
        & \textbf{ReQueR} & \textbf{53.75} & \textbf{36.05} & \textbf{20.30} & \textbf{83.38} & \textbf{87.98} & \textbf{72.00} & \textbf{48.36} & \textbf{57.40}\gain{3.34} \\ 
        \cmidrule[0.25pt](lr){1-10}
        
        % --- Qwen Models ---
        \multicolumn{10}{c}{\textit{\textbf{Qwen Models}}} \\
        \cmidrule[0.25pt](lr){1-10}
        
        \multirow{2}{*}{Qwen3-0.6B} 
        & CoT & 18.44 & 19.14 & 13.14 & 44.48 & 50.80 & 48.80 & 27.53 & 31.76 \\
        & \textbf{ReQueR} & \textbf{23.75} & \textbf{21.66} & \textbf{15.18} & \textbf{61.57} & \textbf{66.14} & \textbf{54.40} & \textbf{30.30} & \textbf{39.00}\gain{7.24} \\ 
        \cmidrule[0.25pt](lr){1-10}
        
        \multirow{2}{*}{Qwen3-1.7B} 
        & CoT & 41.25 & 37.83 & 20.84 & 69.38 & 74.62 & 64.20 & 31.31 & 48.49 \\
        & \textbf{ReQueR} & \textbf{43.13} & \textbf{39.02} & \textbf{22.11} & \textbf{76.71} & \textbf{78.64} & \textbf{77.60} & \textbf{32.83} & \textbf{52.86}\gain{4.37} \\ 
        \cmidrule[0.25pt](lr){1-10}
        
        % \multirow{2}{*}{Qwen3-4B} 
        % & CoT & 63.44 & 48.37 & 26.54 & 81.86 & 84.62 & 83.20 & 40.91 & 61.28 \\
        % & \textbf{ReQueR} & \textbf{64.38} & \textbf{49.41} & \textbf{27.71} & \textbf{83.48} & \textbf{85.38} & \textbf{84.60} & \textbf{43.43} & \textbf{62.63}\gain{1.35} \\ 
        % \cmidrule[0.25pt](lr){1-10}

        \multirow{2}{*}{Qwen3-8B} 
        & CoT & 65.31 & 50.00 & 27.91 & 85.67 & 86.96 & 83.40 & 41.92 & 63.02 \\
        & \textbf{ReQueR} & \textbf{68.13} & \textbf{51.45} & \textbf{28.16} & \textbf{88.05} & \textbf{87.08} & \textbf{85.00} & \textbf{45.45} & \textbf{64.76}\gain{1.74} \\ 
        \cmidrule[0.25pt](lr){1-10}

        \multirow{2}{*}{Qwen2.5-72B-Instruct} 
        & CoT & 61.88 & 43.62 & 26.40 & 81.57 & 82.26 & 76.20 & 50.88 & 60.40 \\
        & \textbf{ReQueR} & \textbf{62.50} & \textbf{46.44} & \textbf{27.37} & \textbf{85.43} & \textbf{88.12} & \textbf{83.00} & \textbf{52.02} & \textbf{63.55}\gain{3.15} \\ 
        \cmidrule[0.25pt](lr){1-10}
        
        % --- Deepseek Models ---
        \multicolumn{10}{c}{\textit{\textbf{Deepseek \& Mixtral Models}}} \\
        \cmidrule[0.25pt](lr){1-10}
        
        \multirow{2}{*}{DS-MoE-16B-Chat} 
        & CoT & 6.25 & 2.52 & 5.94 & 35.24 & 39.48 & 16.20 & 24.75 & 18.63 \\
        & \textbf{ReQueR} & \textbf{7.19} & \textbf{3.56} & \textbf{6.64} & \textbf{45.71} & \textbf{48.34} & \textbf{19.20} & \textbf{26.77} & \textbf{22.49}\gain{3.86} \\ 
        \cmidrule[0.25pt](lr){1-10}

        % % --- Mixtral Models ---
        % \multicolumn{10}{c}{\textit{\textbf{Mixtral Models}}} \\
        % \cmidrule[0.25pt](lr){1-10}
        
        \multirow{2}{*}{Mixtral-8x7B-Instruct} 
        & CoT & 9.69 & 7.12 & 8.63 & 47.29 & 52.12 & 30.60 & 25.76 & 25.89 \\
        & \textbf{ReQueR} & \textbf{18.12} & \textbf{10.68} & \textbf{9.28} & \textbf{57.52} & \textbf{59.66} & \textbf{33.00} & \textbf{31.31} & \textbf{31.37}\gain{5.48} \\ 
        
        \bottomrule
    \end{tabular}
    \vspace{-10pt}
\end{table*}

\section{Experiments}
\label{sec:experiment}

\subsection{Experimental Setup}

\paragraph{Datasets and Benchmarks}
Training data is sourced from GSM8K~\cite{cobbe2021training}, MATH~\cite{hendrycks2021measuring}, and OpenHermes-2.5~\cite{OpenHermes}. We derive 6,144 high-quality samples for cold-start SFT. For the RL phase, the training set consists of 2,048 samples (1,024 each from GSM8K and MATH). To evaluate ReQueR's generalization, we employ diverse reasoning benchmarks: (i) \textbf{Mathematical Reasoning:} GSM-Symbolic~\cite{mirzadeh2024gsm}, GSM-Plus~\cite{li2024gsm}, MATH-500~\cite{lightman2023let}, OlympiadBench~\cite{he2024olympiadbench}, Omni-MATH~\cite{gao2024omni}, and AMC23; (ii) \textbf{General Reasoning:} MMLU-Pro~\cite{wang2024mmlu} and GPQA-Diamond~\cite{rein2024gpqa}.

\paragraph{Refiner and Solver Configurations}
The Refiner $\pi_\theta$ is based on Qwen3-4B. Training is restricted to $\mathcal{S}_{\text{train}} = \{\text{Qwen3-0.6B, 1.7B, 4B}\}$~\cite{yang2025qwen3} in non-thinking mode, while evaluation encompasses a broader $\mathcal{S}_{\text{eval}}$ to verify $O(1)$ transferability. This set features unseen dense architectures (Qwen3-8B, Llama-3.2-3B-Instruct, Llama-3.1-8B/70B-Instruct, Qwen2.5-72B-Instruct)~\cite{yang2025qwen3} and Mixture-of-Experts paradigms (Mixtral-8x7B-Instruct~\cite{jiang2024mixtral}, DeepSeek-MoE-16b-chat~\cite{dai2024deepseekmoe}), testing the meta-policy's robustness across divergent model scales and structural designs.

\paragraph{Baseline Methods}
We benchmark ReQueR against five representative paradigms: 
(1) CoT: Standard CoT prompting; 
(2) Re2~\cite{xu2024re}: Heuristic input re-reading for enhanced encoding; 
(3) RaR~\cite{deng2023rephrase}: Human-defined query rephrasing and expansion; 
(4) TextGrad~\cite{yuksekgonul2024textgrad}: Gradient-based prompt optimization; 
(5) GEPA~\cite{agrawal2025gepa}: Genetic algorithm-based prompt evolution.

\paragraph{Implementation and Metrics}
% The Refiner is optimized via the verl~\cite{sheng2025hybridflow} framework for 100 steps (batch size 64, learning rate $1 \times 10^{-6}$). Performance is measured using the \texttt{Avg@4} accuracy at temperature 0.6. Detailed experimental configurations and hyperparameter settings are provided in Appendix~\ref{appendix:experiment_details}.
The Refiner is optimized via the verl~\cite{sheng2025hybridflow} framework (batch size 64, learning rate $1 \times 10^{-6}$). Performance is measured using the \texttt{Avg@4} accuracy at temperature 0.6. Detailed experimental configurations and hyperparameter settings are provided in Appendix~\ref{appendix:experiment_details}.

\subsection{Main Results}

\begin{table*}[t]
    \centering
    \caption{Performance of ReQueR on Qwen3-1.7B vs. baselines across tasks (best in \textbf{bold}, gains in \textcolor{darkgreen}{green}).}
    \vspace{-6pt}
    \label{tab:main_results2}
    \setlength{\tabcolsep}{3pt} % 调整列间距
    \small % 统一字体大小
    \begin{tabular}{l c c c c c c c}
        \toprule
        \textbf{Method} & \textbf{GSM-Symb.} & \textbf{MATH-500} & \textbf{Omni-MATH} & \textbf{Olym. Bench} & \textbf{GPQA-Diam.} & \textbf{MMLU-Pro} & \textbf{Avg.} \\
        \midrule
        
        Zero-shot CoT & 74.62 & 64.20 & 20.84 & 37.83 & 31.31 & 43.80 & 45.43 \\
        \midrule[0.25pt]
        
        TextGrad      & 76.78 & 72.40 & 20.14 & 34.87 & 31.82 & 44.05 & 46.68 \\
        GEPA          & 76.95 & 73.60 & 20.62 & 35.12 & 32.05 & 45.32 & 47.28 \\
        \midrule[0.25pt]
        
        Re2           & 76.78 & 71.40 & 21.03 & 38.13 & 31.31 & 44.01 & 47.11 \\
        RaR           & 65.64 & 72.80 & 20.37 & 38.58 & 31.82 & 42.98 & 45.37 \\
        \midrule[0.25pt]
        
        \textbf{ReQueR (Ours)} & \textbf{78.64} & \textbf{77.60} & \textbf{22.11} & \textbf{39.02} & \textbf{32.83} & \textbf{46.09} & \textbf{49.38}\gain{2.10} \\
        \bottomrule
    \end{tabular}
    \vspace{-6pt}
\end{table*}

\begin{table*}[t]
    \centering
    \caption{Performance of ReQueR on MATH-500 vs. baselines across model scales (best in \textbf{bold}, gains in \textcolor{darkgreen}{green}).}
    \vspace{-6pt}
    \label{tab:main_results3}
    \setlength{\tabcolsep}{4pt} % 保持较小的列间距以适应宽标题
    \small 
    \begin{tabular}{l c c c c c c c}
        \toprule
        \textbf{Method} & 
        \makecell[c]{\textbf{Llama-3.2-}\\\textbf{3B-Instruct}} & 
        \makecell[c]{\textbf{Llama-3.1-}\\\textbf{8B-Instruct}} & 
        \makecell[c]{\textbf{Llama-3.1-}\\\textbf{70B-Instruct}} & 
        \makecell[c]{\textbf{Qwen3-0.6B}} & 
        \makecell[c]{\textbf{Qwen3-8B}} & 
        \makecell[c]{\textbf{Qwen2.5-}\\\textbf{72B-Instruct}} & 
        \textbf{Avg.} \\
        \midrule
        Zero-shot CoT & 46.60 & 45.80 & 64.60 & 48.80 & 83.40 & 76.20 & 60.90 \\
        \midrule[0.25pt]
        TextGrad      & 43.00 & 48.20 & 68.00 & 46.80 & 83.00 & 82.20 & 61.87 \\
        GEPA          & 44.20 & 49.40 & 68.60 & 47.60 & 83.80 & 82.60 & 62.70 \\
        \midrule[0.25pt]
        Re2           & 46.40 & 50.40 & 67.40 & 48.60 & 84.00 & 82.20 & 63.17 \\
        RaR           & 44.00 & 45.20 & 68.80 & 47.80 & 82.80 & 81.00 & 61.60 \\
        \midrule[0.25pt]
        \textbf{ReQueR (Ours)} & \textbf{50.20} & \textbf{54.00} & \textbf{72.00} & \textbf{54.40} & \textbf{85.00} & \textbf{83.00} & \textbf{66.43}\gain{3.26} \\
        \bottomrule
    \end{tabular}
    \vspace{-10pt}
\end{table*}

\paragraph{Transferability across Models and Tasks}
Table~\ref{tab:main_results1} demonstrates that by training on a limited 2,048-sample subset of GSM8K and MATH using the small-scale $\mathcal{S}_{\text{train}}$, ReQueR achieves consistent performance gains across all seven reasoning benchmarks. The Refiner exhibits robust cross-model transferability, significantly elevating smaller Solvers such as Qwen3-0.6B (+7.24\%) while successfully generalizing to large-scale and heterogeneous architectures, including Llama-3.1-70B-Instruct (+3.34\%) and Qwen2.5-72B-Instruct (+3.15\%). Notably, the Refiner also improves the MoE-based DeepSeek (+3.86\%) and Mixtral (+5.48\%). These results suggest that $\pi_\theta$ captures universal, architectural-agnostic reasoning triggers. Consequently, ReQueR establishes an efficient one-to-many inference-time paradigm for unlocking the latent potential of frozen LLMs without independent model-tuning.

\paragraph{Comparison with Existing Baselines} 
ReQueR demonstrates superior task-level stability and consistent cross-model activation compared to existing paradigms. As shown in Table~\ref{tab:main_results2}, it outperforms the strongest baseline by an average of 2.10\% across all benchmarks, confirming that RL-driven refinement handles diverse logical distributions more robustly than static optimization. Furthermore, Table~\ref{tab:main_results3} validates its steady activation efficacy across various scales, surpassing the next best method by an average of 3.26\% on MATH-500. This consistent gain from 0.6B to 72B proves that the meta-policy effectively elicits the latent potential of frozen Solvers regardless of their size or architecture.

\subsection{Further Analysis}
\begin{figure}[t]
    \centering
    \includegraphics[width=0.8\linewidth]{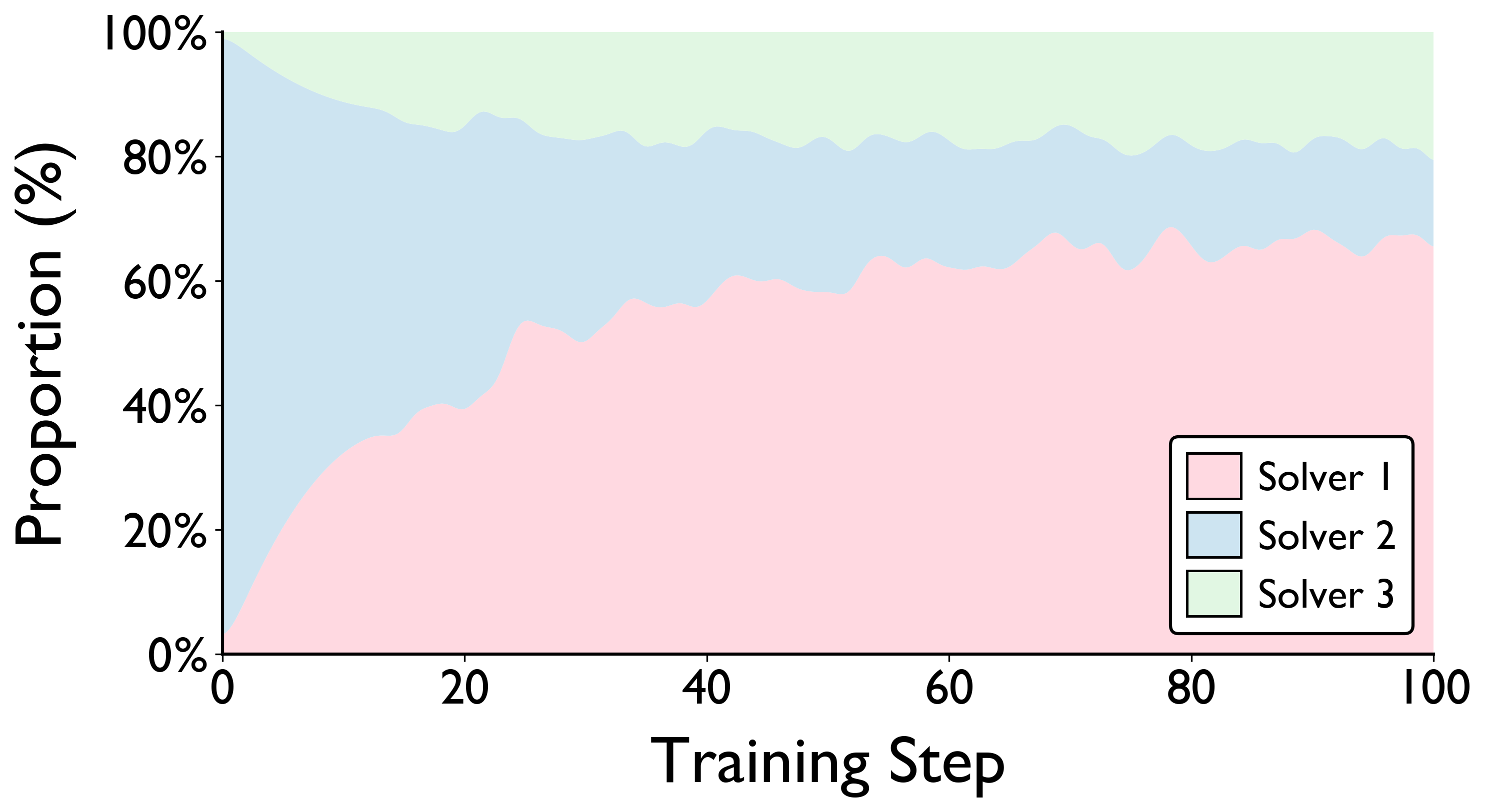}
    \vspace{-5pt}
    \caption{\textbf{Evolution of Solver distribution during training.} ASH dynamically matches Solver difficulty to the Refiner’s competence.} 
    \label{fig:ash_dynamics}
    \vspace{-10pt}
\end{figure}

\begin{figure}[t]
    \centering
    \includegraphics[width=0.8\linewidth]{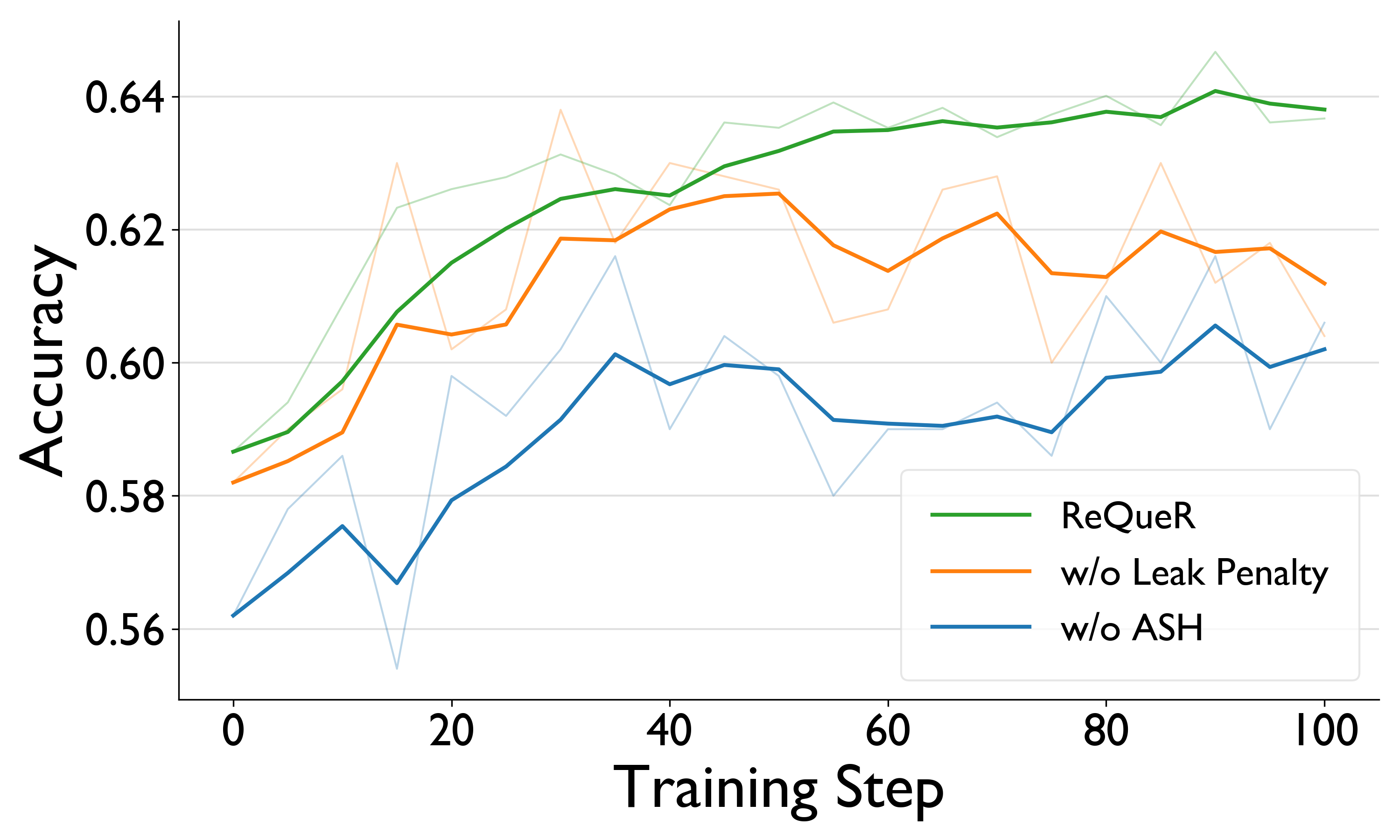}
    \vspace{-5pt}
    \caption{\textbf{Progress of validation accuracy during training.} Both ASH and Leak Penalty are necessary for maintaining stable training and robust generalization.}
    \label{fig:ablation}
    \vspace{-10pt}
\end{figure}

\begin{table}[t]
    \centering
    \caption{Average response length (Tokens / Sample) of Raw queries vs. ReQueR-refined queries.}
    \vspace{-5pt}
    \label{tab:length_delta_analysis}
    \setlength{\tabcolsep}{3pt}
    \small 
    \begin{tabular}{l l c c c}
        \toprule
        \textbf{Model} & \textbf{Task} & \textbf{Raw} & \textbf{ReQueR} & \textbf{$\Delta$ (\%)} \\
        \midrule
        \multirow{3}{*}{\makecell[l]{\textit{Mixtral-8x7B-}\\ \textit{Instruct}}} & GSM8K & 678.0 & 560.8 & 17.3\% $\downarrow$ \\
         & MATH-500 & 1119.2 & 1033.1 & 7.7\% $\downarrow$ \\
         & AMC23 & 11911.5 & 10595.7 & 11.0\% $\downarrow$ \\
        \midrule[0.25pt]
        \multirow{3}{*}{\makecell[l]{\textit{Qwen2.5-72B-}\\ \textit{Instruct}}} & GSM8K & 886.3 & 796.8 & 10.1\% $\downarrow$ \\
         & MATH-500 & 1638.1 & 1559.0 & 4.8\% $\downarrow$ \\
         & AMC23 & 19639.6 & 18045.5 & 8.1\% $\downarrow$ \\
        \bottomrule
    \end{tabular}
    \vspace{-10pt}
\end{table}

\definecolor{MorandiBlue}{RGB}{163, 184, 204}
\definecolor{DeepBlue}{RGB}{70, 90, 110}

\definecolor{MorandiGreen}{RGB}{168, 181, 158}
\definecolor{DeepGreen}{RGB}{80, 95, 70}

\definecolor{MorandiPink}{RGB}{205, 175, 175}
\definecolor{DeepPink}{RGB}{120, 80, 80}

\definecolor{MorandiTan}{RGB}{214, 196, 161}
\definecolor{DeepTan}{RGB}{110, 90, 50}

\newcommand{\stcolor}[2]{
    \setstcolor{#1}
    \st{#2}
}
\makeatletter
\newcommand{\wavecolor}[2]{
    \begingroup
    \def\UL@color{\color{#1}}
    \uwave{\textcolor{black}{#2}}
    \endgroup
}
\makeatother

\newtcolorbox{strategybox}[2]{
    colback=#2!5,
    colframe=#2,
    fonttitle=\small\bfseries,
    title=#1,
    boxrule=0.8pt,
    arc=2pt,
    width=\textwidth,
    left=5pt,
    right=5pt,
    top=1pt,
    bottom=0pt,
    before skip=4pt,
    after skip=0pt
}

\begin{table*}[t]
\centering
\caption{Case studies illustrating how ReQueR transforms raw queries into reasoning-facilitating structures.}
\vspace{-10pt}
\label{tab:prompt_strategies}

% --- Strategy 1: Symbolic Parameterization ---
\begin{strategybox}{Strategy 1: Symbolic Parameterization}{MorandiBlue}
    \scriptsize \textbf{Description:} Transforms narrative word problems into formal symbolic models comprising variables and equations.
    \vspace{2pt} \hrule \vspace{2pt}
    \begin{tabularx}{\textwidth}{l|X|X}
        & \textbf{Raw Query} & \textbf{Refined Query} \\ \hline
        \textbf{Ex.} & Ten treeks weigh as much as three squigs and one goolee. Two treeks and one goolee are equal in weight to one squig. The combined weight of how many treeks equals the weight of one squig? & 
        \textbf{\textcolor{DeepBlue}{Let $T$ be the weight of one treek, $S$ be the weight of one squig, and $G$ be the weight of one goolee.}} We are given: \textbf{\textcolor{DeepBlue}{1. $10T = 3S + G$, 2. $2T + G = S$.}} We need to \textbf{\textcolor{DeepBlue}{find the number $n$ such that $nT = S$}}. Determine $n$ by solving the system of equations. \\
    \end{tabularx}
    \vspace{-2pt}
\end{strategybox}

% --- Strategy 2: Contextual De-noising ---
\begin{strategybox}{Strategy 2: Contextual De-noising}{MorandiGreen}
    \scriptsize \textbf{Description:} Filters out factual but computationally irrelevant narrative noise from the problem description.
    \vspace{2pt} \hrule \vspace{2pt}
    \begin{tabularx}{\textwidth}{l|X|X}
        & \textbf{Raw Query} & \textbf{Refined Query} \\ \hline
        \textbf{Ex.} & \stcolor{DeepGreen}{Daniel works at an electronics store, and he claims that} the popularity of a television is inversely proportional to its cost. If 15 customers buy a television that costs \$1500, how many customers would buy a television that costs \$2500? & 
        \textbf{\textcolor{DeepGreen}{Given that}} the number of customers buying a television is inversely proportional to its cost, and that 15 customers buy a television costing \$1500, how many customers would buy a television costing \$2500? \\
    \end{tabularx}
    \vspace{-2pt}
\end{strategybox}

% --- Strategy 3: Linguistic Disambiguation ---
\begin{strategybox}{Strategy 3: Linguistic Disambiguation}{MorandiPink}
    \scriptsize \textbf{Description:} Resolves semantic ambiguity by replacing vague natural language with formal operators or explicit sets.
    \vspace{2pt} \hrule \vspace{2pt}
    \begin{tabularx}{\textwidth}{l|X|X}
        & \textbf{Raw Query} & \textbf{Refined Query} \\ \hline
        \textbf{Ex.} & What is the \wavecolor{DeepPink}{sum of all of the multiples of 3 between 100 and 200}? & 
        Calculate the sum of the \textbf{\textcolor{DeepPink}{arithmetic series}} starting at \textbf{\textcolor{DeepPink}{102}}, ending at \textbf{\textcolor{DeepPink}{198}}, with a \textbf{\textcolor{DeepPink}{common difference of 3}}. \\
    \end{tabularx}
    \vspace{-2pt}
\end{strategybox}

% --- Strategy 4: Conceptual Reduction ---
\begin{strategybox}{Strategy 4: Conceptual Reduction}{MorandiTan}
    \scriptsize \textbf{Description:} Decomposes high-level math concepts into primitive algebraic operations and decision constraints.
    \vspace{2pt} \hrule \vspace{2pt}
    \begin{tabularx}{\textwidth}{l|X|X}
        & \textbf{Raw Query} & \textbf{Refined Query} \\ \hline
        \textbf{Ex.} & How many \wavecolor{DeepTan}{vertical asymptotes} does the graph of $y = \frac{2}{x^2+x-6}$ have? & 
        Find the number of \textbf{\textcolor{DeepTan}{distinct real numbers $x$}} for which the \textbf{\textcolor{DeepTan}{denominator}} of the function $y = \frac{2}{x^2+x-6}$ \textbf{\textcolor{DeepTan}{is zero}}. \\
    \end{tabularx}
    \vspace{-2pt}
\end{strategybox}
\vspace{-12pt}
\end{table*}

\paragraph{Emergent Refinement Strategies} RL training yields high-value strategies that underpin ReQueR’s effectiveness. As shown in Table~\ref{tab:prompt_strategies}, we summarize four representative archetypes that distill human intuition into machine-executable logic: (i) Symbolic Parameterization; (ii) Contextual De-noising; (iii) Linguistic Disambiguation; and (iv) Conceptual Reduction. These emergent patterns explain ReQueR’s success: they project raw queries onto a logic-dense manifold specifically tailored to trigger the Solver's latent reasoning. Full taxonomy is provided in Appendix~\ref{app:qualitative}.

\paragraph{Analysis of ASH} Figure~\ref{fig:ash_dynamics} reveals that ASH adaptively shifts the Solver distribution from diverse models to weaker ones as Refiner competence improves. This staged curriculum keeps the model within the Zone of Proximal Development to maintain dense rewards. As shown in Figure~\ref{fig:ablation}, removing ASH degrades performance significantly, confirming that balancing task difficulty with current competence is essential for stable optimization.

\paragraph{Analysis of Leak Penalty} Figure~\ref{fig:ablation} demonstrates that omitting the leakage penalty leads to an early performance surge followed by a late-stage decline. This behavior signifies reward hacking where the model exploits non-reasoning shortcuts. Our PPL-based constraint suppresses verbatim answer leakage and forces the Refiner to elicit latent logic, which ultimately yields superior generalization.

\paragraph{Analysis of Inference Overhead}
The computational overhead introduced by ReQueR depends on the deployment scenario. For static datasets, refinement costs are virtually negligible as queries can be pre-processed in a single pass to serve multiple downstream Solvers. In real-time settings, although the 4B Refiner adds a secondary generation step, the overhead is limited to query reformulation, which is significantly shorter than the subsequent reasoning chains. Crucially, this latency is mitigated by improved Solver efficiency. As quantified in Table~\ref{tab:length_delta_analysis}, refined queries yield shorter responses with length reductions between 4.8\% and 17.3\%. This suggests that ReQueR, by clarifying input logic, facilitates more efficient CoT trajectories that partially offset the initial refinement cost.

\subsection{Ablation Study}

\begin{table}[t]
    \centering
    \caption{\textbf{Comprehensive ablation study.} Impact of training stages (SFT, RL) and specific components (ASH, Leak Penalty) across different base models.}
    \vspace{-5pt}
    \label{tab:combined_ablation}
    \setlength{\tabcolsep}{8pt} 
    \small 
    \begin{tabular}{l c c}
        \toprule
        \textbf{Method} & \textbf{MATH-500} & \textbf{GSM-Plus} \\
        \midrule
        \textit{Qwen3-0.6B} & 49.05 & 45.28 \\
        \quad + SFT & 50.68 & 50.42 \\
        \quad \textbf{+ RL (Full ReQueR)} & \textbf{54.40} & \textbf{61.57} \\
        \quad \quad w/o ASH & 52.12 & 54.85 \\
        \quad \quad w/o Leak Penalty & 53.78 & 59.42 \\
        \midrule[0.25pt]
        \textit{Llama-3.1-8B-Instruct} & 46.12 & 67.24 \\
        \quad + SFT & 48.56 & 68.46 \\
        \quad \textbf{+ RL (Full ReQueR)} & \textbf{54.00} & \textbf{71.76} \\
        \quad \quad w/o ASH & 48.15 & 68.32 \\
        \quad \quad w/o Leak Penalty & 53.28 & 70.05 \\
        \midrule[0.25pt]
        \textit{Qwen2.5-72B-Instruct} & 76.68 & 81.75 \\
        \quad + SFT & 78.72 & 82.81 \\
        \quad \textbf{+ RL (Full ReQueR)} & \textbf{83.00} & \textbf{85.43} \\
        \quad \quad w/o ASH & 78.40 & 82.95 \\
        \quad \quad w/o Leak Penalty & 81.15 & 83.82 \\
        \bottomrule
    \end{tabular}
    \vspace{-12pt}
\end{table}

\paragraph{Ablation on SFT and RL Training} Table~\ref{tab:combined_ablation} reveals the cumulative gains from our progressive training pipeline. SFT establishes a foundational alignment, providing a 5.14\% increase for Qwen3-0.6B on GSM-Plus, while subsequent RL further boosts performance by 11.15\%. This trajectory confirms that RL is essential to fully activate the reasoning triggers initialized during SFT.

\paragraph{Ablation on ASH and Leak Penalty} Table~\ref{tab:combined_ablation} identifies ASH as the most vital component for training stability. Its removal leads to a significant 5.85\% drop for Llama-3.1-8B-Instruct on MATH-500, illustrating its role in preserving the optimal difficulty-competence balance. The leak penalty further ensures policy integrity by preventing reward hacking, as evidenced by the 2.15\% decrease on GSM-Plus for Qwen3-0.6B upon its removal.

\section{Conclusion}
In this paper, we presented ReQueR, a unified inference-time Refiner policy to activate the latent reasoning capabilities of diverse, frozen LLMs. By shifting the alignment burden from individual model-tuning to a scalable $O(1)$ paradigm, our framework enables a single Refiner to elicit reasoning without Solver-parameter updates. Empirical results across diverse architectures and scales indicate the robust generalization of ReQueR, highlighting its potential as an efficient one-to-many solution for frozen LLMs. Furthermore, this approach offers significant potential for agentic workflows, such as code agents, where dynamic query refinement can effectively bridge the gap between ambiguous human intent and coding LLMs. These findings highlight ReQueR as a versatile and efficient foundation for future research into adaptive, inference-time reasoning elicitation.

\section*{Limitations}
Despite its efficacy, ReQueR faces several constraints. The two-stage pipeline introduces additional latency and the risk of cascading errors, where hallucinations during refinement may mislead the Solver. Theoretically, our framework is adaptable to diverse task styles (including open-ended generation and creative writing), with the Refiner policy capable of learning specialized strategies. However, formulating rewards for non-verifiable tasks and effectively balancing reward signals across heterogeneous domains remain open challenges, currently confining our application to reasoning tasks with verifiable ground truth. Furthermore, while the ASH curriculum promotes model-agnostic triggers, the Refiner remains influenced by the architectural biases of its training pool. Finally, as an elicitor rather than a generator, ReQueR can only activate a Solver's latent potential and cannot unlock reasoning capabilities fundamentally absent from its pretrained weights.

\section*{Acknowledgments}
This work was supported in part by the National Natural Science Foundation of China (92570101, 62502447).

\bibliography{custom}

@article{wei2022chain,
  title={Chain-of-thought prompting elicits reasoning in large language models},
  author={Wei, Jason and Wang, Xuezhi and Schuurmans, Dale and Bosma, Maarten and Xia, Fei and Chi, Ed and Le, Quoc V and Zhou, Denny and others},
  journal={Advances in neural information processing systems},
  volume={35},
  pages={24824--24837},
  year={2022}
}

@article{shao2024deepseekmath,
  title={Deepseekmath: Pushing the limits of mathematical reasoning in open language models},
  author={Shao, Zhihong and Wang, Peiyi and Zhu, Qihao and Xu, Runxin and Song, Junxiao and Bi, Xiao and Zhang, Haowei and Zhang, Mingchuan and Li, YK and Wu, Yang and others},
  journal={arXiv preprint arXiv:2402.03300},
  year={2024}
}

@inproceedings{zhao2021calibrate,
  title={Calibrate before use: Improving few-shot performance of language models},
  author={Zhao, Zihao and Wallace, Eric and Feng, Shi and Klein, Dan and Singh, Sameer},
  booktitle={International conference on machine learning},
  pages={12697--12706},
  year={2021},
  organization={PMLR}
}

@article{mirzadeh2024gsm,
  title={Gsm-symbolic: Understanding the limitations of mathematical reasoning in large language models},
  author={Mirzadeh, Iman and Alizadeh, Keivan and Shahrokhi, Hooman and Tuzel, Oncel and Bengio, Samy and Farajtabar, Mehrdad},
  journal={arXiv preprint arXiv:2410.05229},
  year={2024}
}

@article{cai2025does,
  title={Does Tone Change the Answer? Evaluating Prompt Politeness Effects on Modern LLMs: GPT, Gemini, LLaMA},
  author={Cai, Hanyu and Shen, Binqi and Jin, Lier and Hu, Lan and Fan, Xiaojing},
  journal={arXiv preprint arXiv:2512.12812},
  year={2025}
}

@article{deng2023rephrase,
  title={Rephrase and respond: Let large language models ask better questions for themselves},
  author={Deng, Yihe and Zhang, Weitong and Chen, Zixiang and Gu, Quanquan},
  journal={arXiv preprint arXiv:2311.04205},
  year={2023}
}

@article{hu2022lora,
  title={Lora: Low-rank adaptation of large language models.},
  author={Hu, Edward J and Shen, Yelong and Wallis, Phillip and Allen-Zhu, Zeyuan and Li, Yuanzhi and Wang, Shean and Wang, Lu and Chen, Weizhu and others},
  journal={ICLR},
  volume={1},
  number={2},
  pages={3},
  year={2022}
}

@article{guo2025deepseek,
  title={Deepseek-r1: Incentivizing reasoning capability in llms via reinforcement learning},
  author={Guo, Daya and Yang, Dejian and Zhang, Haowei and Song, Junxiao and Zhang, Ruoyu and Xu, Runxin and Zhu, Qihao and Ma, Shirong and Wang, Peiyi and Bi, Xiao and others},
  journal={arXiv preprint arXiv:2501.12948},
  year={2025}
}

@inproceedings{yang2023large,
  title={Large language models as optimizers},
  author={Yang, Chengrun and Wang, Xuezhi and Lu, Yifeng and Liu, Hanxiao and Le, Quoc V and Zhou, Denny and Chen, Xinyun},
  booktitle={The Twelfth International Conference on Learning Representations},
  year={2023}
}

@article{yu2025dapo,
  title={Dapo: An open-source llm reinforcement learning system at scale},
  author={Yu, Qiying and Zhang, Zheng and Zhu, Ruofei and Yuan, Yufeng and Zuo, Xiaochen and Yue, Yu and Dai, Weinan and Fan, Tiantian and Liu, Gaohong and Liu, Lingjun and others},
  journal={arXiv preprint arXiv:2503.14476},
  year={2025}
}

@inproceedings{gao2023scaling,
  title={Scaling laws for reward model overoptimization},
  author={Gao, Leo and Schulman, John and Hilton, Jacob},
  booktitle={International Conference on Machine Learning},
  pages={10835--10866},
  year={2023},
  organization={PMLR}
}

@book{vygotsky1978mind,
  title={Mind in society: The development of higher psychological processes},
  author={Vygotsky, Lev S},
  volume={86},
  year={1978},
  publisher={Harvard university press}
}

@inproceedings{bengio2009curriculum,
  title={Curriculum learning},
  author={Bengio, Yoshua and Louradour, J{\'e}r{\^o}me and Collobert, Ronan and Weston, Jason},
  booktitle={Proceedings of the 26th annual international conference on machine learning},
  pages={41--48},
  year={2009}
}

@inproceedings{wang2024large,
  title={Large language models are not fair evaluators},
  author={Wang, Peiyi and Li, Lei and Chen, Liang and Cai, Zefan and Zhu, Dawei and Lin, Binghuai and Cao, Yunbo and Kong, Lingpeng and Liu, Qi and Liu, Tianyu and others},
  booktitle={Proceedings of the 62nd Annual Meeting of the Association for Computational Linguistics (Volume 1: Long Papers)},
  pages={9440--9450},
  year={2024}
}

@article{gulcehre2023reinforced,
  title={Reinforced self-training (rest) for language modeling},
  author={Gulcehre, Caglar and Paine, Tom Le and Srinivasan, Srivatsan and Konyushkova, Ksenia and Weerts, Lotte and Sharma, Abhishek and Siddhant, Aditya and Ahern, Alex and Wang, Miaosen and Gu, Chenjie and others},
  journal={arXiv preprint arXiv:2308.08998},
  year={2023}
}

@article{le2022coderl,
  title={Coderl: Mastering code generation through pretrained models and deep reinforcement learning},
  author={Le, Hung and Wang, Yue and Gotmare, Akhilesh Deepak and Savarese, Silvio and Hoi, Steven Chu Hong},
  journal={Advances in Neural Information Processing Systems},
  volume={35},
  pages={21314--21328},
  year={2022}
}

@article{yuksekgonul2024textgrad,
  title={Textgrad: Automatic" differentiation" via text},
  author={Yuksekgonul, Mert and Bianchi, Federico and Boen, Joseph and Liu, Sheng and Huang, Zhi and Guestrin, Carlos and Zou, James},
  journal={arXiv preprint arXiv:2406.07496},
  year={2024}
}

@article{agrawal2025gepa,
  title={Gepa: Reflective prompt evolution can outperform reinforcement learning},
  author={Agrawal, Lakshya A and Tan, Shangyin and Soylu, Dilara and Ziems, Noah and Khare, Rishi and Opsahl-Ong, Krista and Singhvi, Arnav and Shandilya, Herumb and Ryan, Michael J and Jiang, Meng and others},
  journal={arXiv preprint arXiv:2507.19457},
  year={2025}
}

@inproceedings{sheng2025hybridflow,
  title={Hybridflow: A flexible and efficient rlhf framework},
  author={Sheng, Guangming and Zhang, Chi and Ye, Zilingfeng and Wu, Xibin and Zhang, Wang and Zhang, Ru and Peng, Yanghua and Lin, Haibin and Wu, Chuan},
  booktitle={Proceedings of the Twentieth European Conference on Computer Systems},
  pages={1279--1297},
  year={2025}
}

@inproceedings{xu2024re,
  title={Re-reading improves reasoning in large language models},
  author={Xu, Xiaohan and Tao, Chongyang and Shen, Tao and Xu, Can and Xu, Hongbo and Long, Guodong and Lou, Jian-Guang and Ma, Shuai},
  booktitle={Proceedings of the 2024 Conference on Empirical Methods in Natural Language Processing},
  pages={15549--15575},
  year={2024}
}

@article{cobbe2021training,
  title={Training verifiers to solve math word problems},
  author={Cobbe, Karl and Kosaraju, Vineet and Bavarian, Mohammad and Chen, Mark and Jun, Heewoo and Kaiser, Lukasz and Plappert, Matthias and Tworek, Jerry and Hilton, Jacob and Nakano, Reiichiro and others},
  journal={arXiv preprint arXiv:2110.14168},
  year={2021}
}

@article{hendrycks2021measuring,
  title={Measuring mathematical problem solving with the math dataset},
  author={Hendrycks, Dan and Burns, Collin and Kadavath, Saurav and Arora, Akul and Basart, Steven and Tang, Eric and Song, Dawn and Steinhardt, Jacob},
  journal={arXiv preprint arXiv:2103.03874},
  year={2021}
}

@misc{OpenHermes,
  title = {OpenHermes 2.5: An Open Dataset of Synthetic Data for Generalist LLM Assistants},
  author = {Teknium},
  year = {2023},
  publisher = {HuggingFace}
}

@article{li2024gsm,
  title={Gsm-plus: A comprehensive benchmark for evaluating the robustness of llms as mathematical problem solvers},
  author={Li, Qintong and Cui, Leyang and Zhao, Xueliang and Kong, Lingpeng and Bi, Wei},
  journal={arXiv preprint arXiv:2402.19255},
  year={2024}
}

@inproceedings{lightman2023let,
  title={Let's verify step by step},
  author={Lightman, Hunter and Kosaraju, Vineet and Burda, Yuri and Edwards, Harrison and Baker, Bowen and Lee, Teddy and Leike, Jan and Schulman, John and Sutskever, Ilya and Cobbe, Karl},
  booktitle={The Twelfth International Conference on Learning Representations},
  year={2023}
}

@inproceedings{he2024olympiadbench,
  title={Olympiadbench: A challenging benchmark for promoting agi with olympiad-level bilingual multimodal scientific problems},
  author={He, Chaoqun and Luo, Renjie and Bai, Yuzhuo and Hu, Shengding and Thai, Zhen and Shen, Junhao and Hu, Jinyi and Han, Xu and Huang, Yujie and Zhang, Yuxiang and others},
  booktitle={Proceedings of the 62nd Annual Meeting of the Association for Computational Linguistics (Volume 1: Long Papers)},
  pages={3828--3850},
  year={2024}
}

@article{gao2024omni,
  title={Omni-math: A universal olympiad level mathematic benchmark for large language models},
  author={Gao, Bofei and Song, Feifan and Yang, Zhe and Cai, Zefan and Miao, Yibo and Dong, Qingxiu and Li, Lei and Ma, Chenghao and Chen, Liang and Xu, Runxin and others},
  journal={arXiv preprint arXiv:2410.07985},
  year={2024}
}

@article{wang2024mmlu,
  title={Mmlu-pro: A more robust and challenging multi-task language understanding benchmark},
  author={Wang, Yubo and Ma, Xueguang and Zhang, Ge and Ni, Yuansheng and Chandra, Abhranil and Guo, Shiguang and Ren, Weiming and Arulraj, Aaran and He, Xuan and Jiang, Ziyan and others},
  journal={Advances in Neural Information Processing Systems},
  volume={37},
  pages={95266--95290},
  year={2024}
}

@inproceedings{rein2024gpqa,
  title={Gpqa: A graduate-level google-proof q\&a benchmark},
  author={Rein, David and Hou, Betty Li and Stickland, Asa Cooper and Petty, Jackson and Pang, Richard Yuanzhe and Dirani, Julien and Michael, Julian and Bowman, Samuel R},
  booktitle={First Conference on Language Modeling},
  year={2024}
}

@article{yang2025qwen3,
  title={Qwen3 technical report},
  author={Yang, An and Li, Anfeng and Yang, Baosong and Zhang, Beichen and Hui, Binyuan and Zheng, Bo and Yu, Bowen and Gao, Chang and Huang, Chengen and Lv, Chenxu and others},
  journal={arXiv preprint arXiv:2505.09388},
  year={2025}
}

@article{jiang2024mixtral,
  title={Mixtral of experts},
  author={Jiang, Albert Q and Sablayrolles, Alexandre and Roux, Antoine and Mensch, Arthur and Savary, Blanche and Bamford, Chris and Chaplot, Devendra Singh and Casas, Diego de las and Hanna, Emma Bou and Bressand, Florian and others},
  journal={arXiv preprint arXiv:2401.04088},
  year={2024}
}

@article{dai2024deepseekmoe,
  title={Deepseekmoe: Towards ultimate expert specialization in mixture-of-experts language models},
  author={Dai, Damai and Deng, Chengqi and Zhao, Chenggang and Xu, RX and Gao, Huazuo and Chen, Deli and Li, Jiashi and Zeng, Wangding and Yu, Xingkai and Wu, Yu and others},
  journal={arXiv preprint arXiv:2401.06066},
  year={2024}
}

@article{rao2025dynamic,
  title={Dynamic Sampling that Adapts: Iterative DPO for Self-Aware Mathematical Reasoning},
  author={Rao, Jun and Liu, Xuebo and Deng, Hexuan and Lin, Zepeng and Yu, Zixiong and Wei, Jiansheng and Meng, Xiaojun and Zhang, Min},
  journal={arXiv preprint arXiv:2505.16176},
  year={2025}
}

@article{fang2026proximity,
  title={Proximity-Based Multi-Turn Optimization: Practical Credit Assignment for LLM Agent Training},
  author={Fang, Yangyi and Lin, Jiaye and Fu, Xiaoliang and Qin, Cong and Shi, Haolin and Liu, Chang and Zhao, Peilin},
  journal={arXiv preprint arXiv:2602.19225},
  year={2026}
}

@article{fang2026allocate,
  title={How to Allocate, How to Learn? Dynamic Rollout Allocation and Advantage Modulation for Policy Optimization},
  author={Fang, Yangyi and Lin, Jiaye and Fu, Xiaoliang and Qin, Cong and Shi, Haolin and Hu, Chaowen and Pan, Lu and Zeng, Ke and Cai, Xunliang},
  journal={arXiv preprint arXiv:2602.19208},
  year={2026}
}

@article{dai2025psyche,
  title={Psyche-r1: Towards reliable psychological llms through unified empathy, expertise, and reasoning},
  author={Dai, Chongyuan and Hu, Jinpeng and Shi, Hongchang and Li, Zhuo and Yang, Xun and Wang, Meng},
  journal={arXiv preprint arXiv:2508.10848},
  year={2025}
}

@article{guo2026rethinking,
  title={Rethinking Table Pruning in TableQA: From Sequential Revisions to Gold Trajectory-Supervised Parallel Search},
  author={Guo, Yu and Ye, Shenghao and Chen, Shuangwu and Wen, Zijian and Zhang, Tao and Bai, Qirui and Jin, Dong and Hou, Yunpeng and He, Huasen and Yang, Jian and others},
  journal={arXiv preprint arXiv:2601.03851},
  year={2026}
}

@article{ye2025tableqa,
  title={When tableqa meets noise: A dual denoising framework for complex questions and large-scale tables},
  author={Ye, Shenghao and Guo, Yu and Jin, Dong and Shen, Yikai and Hou, Yunpeng and Chen, Shuangwu and Yang, Jian and Jiang, Xiaofeng},
  journal={arXiv preprint arXiv:2509.17680},
  year={2025}
}

@inproceedings{lin2026cec,
  title={Cec-zero: Zero-supervision character error correction with self-generated rewards},
  author={Lin, Zhiming and Zhao, Kai and Zhang, Sophie and Yu, Peilai and Xiao, Canran},
  booktitle={Proceedings of the AAAI Conference on Artificial Intelligence},
  volume={40},
  number={28},
  pages={23612--23620},
  year={2026}
}

@inproceedings{xiao2026points,
  title={From points to coalitions: hierarchical contrastive shapley values for prioritizing data samples},
  author={Xiao, Canran and Dou, Jiabao and Lin, Zhiming and Ke, Zong and Hou, Liwei},
  booktitle={Proceedings of the AAAI Conference on Artificial Intelligence},
  volume={40},
  number={19},
  pages={15995--16003},
  year={2026}
}

@inproceedings{xiaometa,
  title={Meta-UCF: Unified Task-Conditioned LoRA Generation for Continual Learning in Large Language Models},
  author={Xiao, ShiLin and Xu, Tianxiang and Xiao, Canran and Luo, Weihao and Hou, Liwei and Zhao, Chuangxin},
  booktitle={The Fourteenth International Conference on Learning Representations},
  year={2026}
}

@article{yao2024swift,
  title={Swift sampler: Efficient learning of sampler by 10 parameters},
  author={Yao, Jiawei and Li, Chuming and Xiao, Canran},
  journal={Advances in Neural Information Processing Systems},
  volume={37},
  pages={59030--59053},
  year={2024}
}

@inproceedings{ma2026gor,
  title={Gor: A unified and extensible generative framework for ordinal regression},
  author={Ma, Hongxu and Zhou, Han and Tian, Kai and Zhang, Xuefeng and Chen, Chunjie and Li, Han and Guan, Jihong and Zhou, Shuigeng},
  booktitle={The Fourteenth International Conference on Learning Representations},
  year={2026}
}

@inproceedings{jin2026invariant,
  title={Invariant Feature Learning for Counterfactual Watch-time Prediction in Video Recommendation},
  author={Jin, Chenghou and Ren, Yixin and Ma, Hongxu and Xia, Yewei and Guan, Yi and Zhang, Hao and Ding, Jiandong and Guan, Jihong and Zhou, Shuigeng},
  booktitle={Proceedings of the AAAI Conference on Artificial Intelligence},
  volume={40},
  number={17},
  pages={14964--14972},
  year={2026}
}

@article{miao2026seeing,
  title={Seeing with You: Perception-Reasoning Coevolution for Multimodal Reasoning},
  author={Miao, Ziqi and Jia, Haonan and Li, Lijun and Qian, Chen and Xiong, Yuan and Yan, Wenting and Shao, Jing},
  journal={arXiv preprint arXiv:2603.28618},
  year={2026}
}

@article{yu2026unveiling,
  title={Unveiling Implicit Advantage Symmetry: Why GRPO Struggles with Exploration and Difficulty Adaptation},
  author={Yu, Zhiqi and Chen, Zhangquan and Liu, Mengting and Zhang, Heye and Qu, Liangqiong},
  journal={arXiv preprint arXiv:2602.05548},
  year={2026}
}

@article{ding2026prpo,
  title={PRPO: Aligning Process Reward with Outcome Reward in Policy Optimization},
  author={Ding, Ruiyi and Lv, Yongxuan and Meng, Xianhui and Song, Jiahe and Wang, Chao and Jiang, Chen and Cheng, Yuan},
  journal={arXiv preprint arXiv:2601.07182},
  year={2026}
}

@article{liu2025uniform,
  title={From Uniform to Heterogeneous: Tailoring Policy Optimization to Every Token's Nature},
  author={Liu, Zheng and Liu, Mengjie and Wen, Siwei and Cai, Mengzhang and Cui, Bin and He, Conghui and Zhang, Wentao},
  journal={arXiv preprint arXiv:2509.16591},
  year={2025}
}

@article{xie2025unlocking,
  title={Unlocking exploration in rlvr: Uncertainty-aware advantage shaping for deeper reasoning},
  author={Xie, Can and Pan, Ruotong and Wu, Xiangyu and Zhang, Yunfei and Fu, Jiayi and Gao, Tingting and Zhou, Guorui},
  journal={arXiv preprint arXiv:2510.10649},
  year={2025}
}

@article{li2025curriculum,
  title={Curriculum-rlaif: Curriculum alignment with reinforcement learning from ai feedback},
  author={Li, Mengdi and Lin, Jiaye and Zhao, Xufeng and Lu, Wenhao and Zhao, Peilin and Wermter, Stefan and Wang, Di},
  journal={arXiv preprint arXiv:2505.20075},
  year={2025}
}

@article{lin2025se,
  title={Se-agent: Self-evolution trajectory optimization in multi-step reasoning with llm-based agents},
  author={Lin, Jiaye and Guo, Yifu and Han, Yuzhen and Hu, Sen and Ni, Ziyi and Wang, Licheng and Chen, Mingguang and Liu, Hongzhang and Chen, Ronghao and He, Yangfan and others},
  journal={arXiv preprint arXiv:2508.02085},
  year={2025}
}

@article{wu2025table,
  title={Table-r1: Region-based reinforcement learning for table understanding},
  author={Wu, Zhenhe and Yang, Jian and Liu, Jiaheng and Wu, Xianjie and Pan, Changzai and Zhang, Jie and Zhao, Yu and Song, Shuangyong and Li, Yongxiang and Li, Zhoujun},
  journal={arXiv preprint arXiv:2505.12415},
  year={2025}
}

@article{ling2026neural,
  title={Neural Chain-of-Thought Search: Searching the Optimal Reasoning Path to Enhance Large Language Models},
  author={Ling, Guoming and Huang, Zhongzhan and Lin, Yupei and Li, Junxin and Zhong, Shanshan and Wu, Hefeng and Lin, Liang},
  journal={arXiv preprint arXiv:2601.11340},
  year={2026}
}

@article{zhang2026logical,
  title={Logical Phase Transitions: Understanding Collapse in LLM Logical Reasoning},
  author={Zhang, Xinglang and Zhang, Yunyao and Chen, ZeLiang and Yu, Junqing and Yang, Wei and Song, Zikai},
  journal={arXiv preprint arXiv:2601.02902},
  year={2026}
}

@article{wu2026step,
  title={Step Potential Advantage Estimation: Harnessing Intermediate Confidence and Correctness for Efficient Mathematical Reasoning},
  author={Wu, Fei and Zhang, Zhenrong and Chang, Qikai and Zhang, Jianshu and Liu, Quan and Du, Jun},
  journal={arXiv preprint arXiv:2601.03823},
  year={2026}
}

@inproceedings{zhu2026medeyes,
  title={Medeyes: Learning dynamic visual focus for medical progressive diagnosis},
  author={Zhu, Chunzheng and Lin, Yangfang and Chen, Shen and Wang, Yijun and Lin, Jianxin},
  booktitle={Proceedings of the AAAI Conference on Artificial Intelligence},
  volume={40},
  number={16},
  pages={13916--13924},
  year={2026}
}

@inproceedings{zhu2025pathology,
  title={Pathology-Aware Prototype Evolution via LLM-Driven Semantic Disambiguation for Multicenter Diabetic Retinopathy Diagnosis},
  author={Zhu, Chunzheng and Lin, Yangfang and Shao, Jialin and Lin, Jianxin and Wang, Yijun},
  booktitle={Proceedings of the 33rd ACM International Conference on Multimedia},
  pages={9196--9205},
  year={2025}
}

@article{hao2025rethinking,
  title={Rethinking entropy interventions in rlvr: An entropy change perspective},
  author={Hao, Zhezheng and Wang, Hong and Liu, Haoyang and Luo, Jian and Yu, Jiarui and Dong, Hande and Lin, Qiang and Wang, Can and Chen, Jiawei},
  journal={arXiv preprint arXiv:2510.10150},
  year={2025}
}

@inproceedings{xiao2025visual,
  title={Visual instance-aware prompt tuning},
  author={Xiao, Xi and Zhang, Yunbei and Li, Xingjian and Wang, Tianyang and Wang, Xiao and Wei, Yuxiang and Hamm, Jihun and Xu, Min},
  booktitle={Proceedings of the 33rd ACM International Conference on Multimedia},
  pages={2880--2889},
  year={2025}
}

@article{zhao2026reinforced,
  title={Reinforced Efficient Reasoning via Semantically Diverse Exploration},
  author={Zhao, Ziqi and Ren, Zhaochun and Zou, Jiahong and Yang, Liu and Xu, Zhiwei and Ge, Xuri and Chen, Zhumin and Ma, Xinyu and Shi, Daiting and Wang, Shuaiqiang and others},
  journal={arXiv preprint arXiv:2601.05053},
  year={2026}
}

@article{an2025amo,
  title={Amo-bench: Large language models still struggle in high school math competitions},
  author={An, Shengnan and Cai, Xunliang and Cao, Xuezhi and Li, Xiaoyu and Lin, Yehao and Liu, Junlin and Lv, Xinxuan and Ma, Dan and Wang, Xuanlin and Wang, Ziwen and others},
  journal={arXiv preprint arXiv:2510.26768},
  year={2025}
}

@article{zhou2026look,
  title={Look inward to explore outward: Learning temperature policy from llm internal states via hierarchical rl},
  author={Zhou, Yixiao and Li, Yang and Cheng, Dongzhou and Fan, Hehe and Cheng, Yu},
  journal={arXiv preprint arXiv:2602.13035},
  year={2026}
}

@inproceedings{chang2026balora,
  title={{BA}-Lo{RA}: Bias-Alleviating Low-Rank Adaptation to Mitigate Catastrophic Inheritance in Large Language Models},
  author={Chang, Yupeng and Chang, Yi and Wu, Yuan},
  booktitle={The Fourteenth International Conference on Learning Representations},
  year={2026}
}

@article{li2025choice,
  title={The choice of divergence: A neglected key to mitigating diversity collapse in reinforcement learning with verifiable reward},
  author={Li, Long and Zhou, Zhijian and Hao, Jiaran and Liu, Jason Klein and Miao, Yanting and Pang, Wei and Tan, Xiaoyu and Chu, Wei and Wang, Zhe and Pan, Shirui and others},
  journal={arXiv preprint arXiv:2509.07430},
  year={2025}
}

@article{dong2025aurora,
  title={Aurora: Breaking low-rank bottleneck of lora with nonlinear mapping},
  author={Dong, Haonan and Zhu, Wenhao and Song, Guojie and Wang, Liang},
  journal={arXiv preprint arXiv:2505.18738},
  year={2025}
}

@article{dong2026neureasoner,
  title={NeuReasoner: Towards Explainable, Controllable, and Unified Reasoning via Mixture-of-Neurons},
  author={Dong, Haonan and Jiang, Kehan and Ye, Haoran and Zhu, Wenhao and Kang, Zhaolu and Song, Guojie},
  journal={arXiv preprint arXiv:2604.02972},
  year={2026}
}

@article{jiang2026foe,
  title={FoE: Forest of Errors Makes the First Solution the Best in Large Reasoning Models},
  author={Jiang, Kehan and Dong, Haonan and Kang, Zhaolu and Zhu, Zhengzhou and Song, Guojie},
  journal={arXiv preprint arXiv:2604.02967},
  year={2026}
}

@article{ren2025riskpo,
  title={RiskPO: Risk-based Policy Optimization via Verifiable Reward for LLM Post-Training},
  author={Ren, Tao and Jiang, Jinyang and Yang, Hui and Tian, Wan and Zou, Minhao and Li, Guanghao and Zhang, Zishi and Wang, Qinghao and Qin, Shentao and Zhao, Yanjun and others},
  journal={arXiv preprint arXiv:2510.00911},
  year={2025}
}

@inproceedings{zhao2026saber,
  title={Saber: Switchable and balanced training for efficient llm reasoning},
  author={Zhao, Kai and Zhao, Yanjun and Song, Jiaming and He, Shien and Zhang, Lusheng and Zhang, Qiang and Li, Tianjiao},
  booktitle={Proceedings of the AAAI Conference on Artificial Intelligence},
  volume={40},
  number={41},
  pages={34950--34958},
  year={2026}
}

@inproceedings{li2024towards,
  title={Towards effective data-free knowledge distillation via diverse diffusion augmentation},
  author={Li, Muquan and Zhang, Dongyang and He, Tao and Xie, Xiurui and Li, Yuan-Fang and Qin, Ke},
  booktitle={Proceedings of the 32nd ACM International Conference on Multimedia},
  pages={4416--4425},
  year={2024}
}

@article{dou2025plan,
  title={Plan Then Action: High-Level Planning Guidance Reinforcement Learning for LLM Reasoning},
  author={Dou, Zhihao and Zhao, Qinjian and Wan, Zhongwei and Zhang, Dinggen and Wang, Weida and Raiyan, Towsif and Chen, Benteng and Pan, Qingtao and Ouyang, Yang and Gao, Zhiqiang and others},
  journal={arXiv preprint arXiv:2510.01833},
  year={2025}
}

@inproceedings{chen2024self,
  title={Self-Evolution Fine-Tuning for Policy Optimization},
  author={Chen, Ruijun and Liang, Jiehao and Gao, Shiping and Wan, Fanqi and Quan, Xiaojun},
  booktitle={Findings of the Association for Computational Linguistics: EMNLP 2024},
  pages={4120--4137},
  year={2024}
}

@article{li2026dyjr,
  title={DyJR: Preserving Diversity in Reinforcement Learning with Verifiable Rewards via Dynamic Jensen-Shannon Replay},
  author={Li, Long and Zhou, Zhijian and Wang, Tianyi and Xu, Weidi and Huang, Zuming and Chu, Wei and Wang, Zhe and Pan, Shirui and Qu, Chao and Qi, Yuan},
  journal={arXiv preprint arXiv:2603.16157},
  year={2026}
}

\appendix

\section{Paradigm Comparison: ReQueR vs. Prompt Optimization}
\label{appendix:paradigm_comparison}

To further clarify the novelty of ReQueR, we formalize the paradigmatic divergence between our proposed instance-level refinement and Automatic Prompt Optimization (APO).

\subsection{APO as a Static Task-level Wrapper}
APO methods (e.g., OPRO, Textgrad, GEPA) treat the prompt $P$ as a discrete, static hyperparameter tailored for a specific model $M_i$ on a task distribution $\mathcal{D}_k$. This approach can be formulated as a point-wise search for a global string:
\begin{equation}
\begin{aligned}
& \hat{P}_{\text{APO}} = \arg\max_{P} \mathbb{E}_{x \sim \mathcal{D}_k} \Big[ \\
& \quad \mathcal{R} \left( M_i ( P \oplus x ), y^* \right) \Big],
\end{aligned}
\end{equation}
where $\oplus$ denotes string concatenation. Here, $P$ acts as a fixed wrapper, a ``one-size-fits-all'' instruction prefix. Because this wrapper is tightly coupled to model $M_i$'s internal biases, it lacks the flexibility to adapt to per-sample nuances, leading to an $O(N)$ migration complexity where $N$ is the number of target Solvers.

\subsection{ReQueR as an Instance-level Meta-Policy}
In contrast, ReQueR optimizes a generative policy $\pi_\theta$ that produces a unique refinement $x'$ for every unique input $x$. By optimizing against the Solver manifold $\mathcal{S}$ via the ASH hierarchy, the Refiner learns a universal mapping:
\begin{equation}
\begin{aligned}
& \theta^* = \arg\max_{\theta} \mathbb{E}_{x \sim \mathcal{D}} \mathbb{E}_{s \sim \mathcal{S}} \mathbb{E}_{x' \sim \pi_\theta(x'|x)} \Big[ \\
& \quad \mathcal{R} \left( s(x'), y^* \right) \Big].
\end{aligned}
\end{equation}
Unlike APO, the optimization target here is a \textbf{Meta-refinement Policy}. This shift yields several critical advantages:
\begin{itemize}
    \item \textbf{Per-sample Elasticity:} For every input $x$, the Refiner generates a tailored $x'$ that rectifies the specific logical vulnerabilities of that sample, acting as a dynamic transformation rather than a static bandage. Related per-sample strategies have been explored in visual prompt tuning~\cite{xiao2025visual} and SFT data construction~\cite{chen2024self}.
    \item \textbf{Model-Agnostic Invariance:} Since the policy is trained against the distribution $\mathcal{S}$, it internalizes a meta-strategy for query clarification that is effective across all LLM architectures, enabling $O(1)$ universal transfer.
    \item \textbf{Inference-time Practicality and Robustness:} A significant limitation of APO is its dependence on a priori knowledge of the model and task category during inference. In real-world scenarios, one cannot assume the specific Solver identity or that a query belongs to a singular, predefined task distribution. APO's static wrappers often fail when encountering hybrid-style queries or out-of-distribution (OOD) inputs. In contrast, ReQueR's meta-policy treats every input as a fresh structural disambiguation task, ensuring high-fidelity reasoning elicitation regardless of the underlying task category or Solver backend.
\end{itemize}

\begin{table*}[t]
    \centering
    \caption{Conceptual Comparison: APO vs. ReQueR.}
    \label{tab:paradigm_diff}
    \small
    \renewcommand{\arraystretch}{1.3}
    \begin{tabularx}{\textwidth}{l X X}
        \toprule
        \textbf{Dimension} & \textbf{Automatic Prompt Optimization} & \textbf{ReQueR (Meta-Policy)} \\
        \midrule
        \textbf{Optimization Goal} & Best wrapper string for a whole task & \textbf{Meta-refinement policy} for every input $x$ \\
        \textbf{Granularity} & Global / Dataset-level & Local / Instance-level \\
        \textbf{Mechanism} & Static Prefixing ($P \oplus x$) & Dynamic Transformation ($f: x \to x'$) \\
        % \textbf{Complexity} & $O(N)$ (Highly Model-coupled) & $O(1)$ (Model-agnostic Interface) \\
        \bottomrule
    \end{tabularx}
\end{table*}

The core conceptual differences are summarized in Table~\ref{tab:paradigm_diff}.

\section{Extended Related Work on RL for Reasoning}
\label{appendix:extended_rl}

Beyond the core RL-for-reasoning line discussed in the main paper, recent advances have explored several orthogonal directions. On reward and advantage design, a range of studies investigate dynamic sampling, credit assignment, step-level advantage estimation, and risk-aware objectives to produce denser and more informative training signals~\cite{rao2025dynamic,fang2026proximity,fang2026allocate,wu2026step,ren2025riskpo}. For exploration and efficiency, works target token-level policy tailoring, uncertainty-aware advantage shaping, diversity-promoting exploration, inference-time temperature control, balanced training pipelines, and planning-guided RL~\cite{liu2025uniform,xie2025unlocking,zhao2026reinforced,zhou2026look,zhao2026saber,dou2025plan}. Another line extends RL-for-reasoning beyond pure text, covering multimodal perception-reasoning coevolution and neural chain-of-thought search~\cite{miao2026seeing,ling2026neural}. Finally, domain-specific applications demonstrate the broader impact of RL-based reasoning in psychological, tabular, and medical settings~\cite{dai2025psyche,guo2026rethinking,ye2025tableqa,wu2025table,zhu2026medeyes,zhu2025pathology}. These works collectively enrich the RL-for-reasoning landscape and motivate ReQueR's orthogonal focus on inference-time, model-agnostic reasoning elicitation.

\section{Experimental Details and Hyperparameters}
\label{appendix:experiment_details}
\subsection{Model Specifications and Sources}
To ensure transparency and reproducibility, we provide the detailed configurations and official sources of the models utilized in the ReQueR framework. Table~\ref{tab:models} summarizes the Refiner and the heterogeneous Solver pool, highlighting the diversity in architecture and scale that underpins our one-to-many generalization claims.

% \begin{table*}[t]
%     \centering
%     \caption{Summary of Model Architectures and Official Sources.}
%     \label{tab:models}
%     \small
%     \setlength{\tabcolsep}{8pt}
%     \begin{tabular}{l l c l}
%         \toprule
%         \textbf{Role} & \textbf{Model Series} & \textbf{Parameters} & \textbf{Official Source / Hugging Face Link} \\
%         \midrule
%         \textbf{Refiner} & Qwen3 & 4B & \href{https://huggingface.co/Qwen/Qwen3-4B-Instruct}{Qwen/Qwen3-4B} \\
%         \midrule
%         \textbf{Solvers} & Qwen3 (non-thinking) & 0.6B & \href{https://huggingface.co/Qwen/Qwen3-0.6B-Instruct}{Qwen/Qwen3-0.6B} \\
%         & Qwen3 (non-thinking) & 1.7B & \href{https://huggingface.co/Qwen/Qwen3-1.7B-Instruct}{Qwen/Qwen3-1.7B} \\
%         & Qwen3 (non-thinking) & 4B & \href{https://huggingface.co/Qwen/Qwen3-4B-Instruct}{Qwen/Qwen3-4B} \\
%         & Qwen3 (non-thinking) & 8B & \href{https://huggingface.co/Qwen/Qwen3-8B-Instruct}{Qwen/Qwen3-8B} \\
%         & Qwen2.5-Instruct & 72B & \href{https://huggingface.co/Qwen/Qwen2.5-72B-Instruct}{Qwen/Qwen2.5-72B-Instruct} \\
%         \cmidrule{2-4}
%         & Llama-3.2-Instruct & 3B & \href{https://huggingface.co/meta-llama/Llama-3.2-3B-Instruct}{meta-llama/Llama-3.2-3B-Instruct} \\
%         & Llama-3.1-Instruct & 8B & \href{https://huggingface.co/meta-llama/Llama-3.1-8B-Instruct}{meta-llama/Llama-3.1-8B-Instruct} \\
%         & Llama-3.1-Instruct & 70B & \href{https://huggingface.co/meta-llama/Llama-3.1-70B-Instruct}{meta-llama/Llama-3.1-70B-Instruct} \\
%         \cmidrule{2-4}
%         & DeepSeek-MoE-Chat & 16B & \href{https://huggingface.co/deepseek-ai/deepseek-moe-16b-chat}{deepseek-ai/deepseek-moe-16b-chat} \\
%         & Mixtral-8x7B-Instruct & 47B & \href{https://huggingface.co/mistralai/Mixtral-8x7B-v0.1}{mistralai/Mixtral-8x7B-v0.1} \\
%         \bottomrule
%     \end{tabular}
% \end{table*}
\begin{table*}[t]
    \centering
    \caption{Summary of Model Architectures and Official Sources.}
    \label{tab:models}
    \small
    \setlength{\tabcolsep}{8pt} % 调节列间距
    \begin{tabular}{l l l}
        \toprule
        \textbf{Role} & \textbf{Model} & \textbf{Official Source / Hugging Face Link} \\
        \midrule
        \textbf{Refiner} & Qwen3-4B & \href{https://huggingface.co/Qwen/Qwen3-4B}{Qwen/Qwen3-4B} \\
        \midrule
        \textbf{Solvers} & Qwen3-0.6B$^\ast$ & \href{https://huggingface.co/Qwen/Qwen3-0.6B}{Qwen/Qwen3-0.6B} \\
        & Qwen3-1.7B$^\ast$ & \href{https://huggingface.co/Qwen/Qwen3-1.7B}{Qwen/Qwen3-1.7B} \\
        & Qwen3-4B$^\ast$ & \href{https://huggingface.co/Qwen/Qwen3-4B}{Qwen/Qwen3-4B} \\
        & Qwen3-8B$^\ast$ & \href{https://huggingface.co/Qwen/Qwen3-8B}{Qwen/Qwen3-8B} \\
        & Qwen2.5-72B-Instruct & \href{https://huggingface.co/Qwen/Qwen2.5-72B-Instruct}{Qwen/Qwen2.5-72B-Instruct} \\
        \cmidrule{2-3}
        & Llama-3.2-3B-Instruct & \href{https://huggingface.co/meta-llama/Llama-3.2-3B-Instruct}{meta-llama/Llama-3.2-3B-Instruct} \\
        & Llama-3.1-8B-Instruct & \href{https://huggingface.co/meta-llama/Llama-3.1-8B-Instruct}{meta-llama/Llama-3.1-8B-Instruct} \\
        & Llama-3.1-70B-Instruct & \href{https://huggingface.co/meta-llama/Llama-3.1-70B-Instruct}{meta-llama/Llama-3.1-70B-Instruct} \\
        \cmidrule{2-3}
        & DeepSeek-MoE-16B-Chat & \href{https://huggingface.co/deepseek-ai/deepseek-moe-16b-chat}{deepseek-ai/deepseek-moe-16b-chat} \\
        & Mixtral-8x7B-Instruct & \href{https://huggingface.co/mistralai/Mixtral-8x7B-v0.1}{mistralai/Mixtral-8x7B-v0.1} \\
        \bottomrule
        \multicolumn{3}{l}{\scriptsize $^\ast$ Non-thinking mode (To evaluate the direct reasoning elicitation effect without internal CoT buffers.).}
    \end{tabular}
\end{table*}

\subsection{Dataset Specifications and Benchmarks}
To evaluate the reasoning elicitation and generalization capabilities of ReQueR, we utilize a comprehensive set of datasets for both policy training and multi-domain evaluation. Table~\ref{tab:datasets} summarizes the scope, specific tasks, and official sources for each dataset used in our experiments.

\begin{table*}[t]
    \centering
    \caption{Summary of Datasets for Training and Evaluation.}
    \label{tab:datasets}
    \small
    \setlength{\tabcolsep}{6pt}
    \begin{tabular}{l l l l}
        \toprule
        \textbf{Phase} & \textbf{Dataset} & \textbf{Samples / Scope} & \textbf{Official Source / Hugging Face Link} \\
        \midrule
        \textbf{Training} & OpenHermes-2.5~\cite{OpenHermes} & 6,144 (SFT) & \href{https://huggingface.co/datasets/teknium/OpenHermes-2.5}{teknium/OpenHermes-2.5} \\
        & GSM8K~\cite{cobbe2021training} & 1,024 (RL) & \href{https://huggingface.co/datasets/openai/gsm8k}{openai/gsm8k} \\
        & MATH~\cite{hendrycks2021measuring} & 1,024 (RL) & \href{https://huggingface.co/datasets/hendrycks/competition_math}{hendrycks/competition\_math} \\
        \midrule
        \textbf{Evaluation} & MATH-500~\cite{lightman2023let} & 500 (Hard Math) & \href{https://huggingface.co/datasets/HuggingFaceH4/MATH-500}{HuggingFaceH4/MATH-500} \\
        (Mathematical) & GSM-Symbolic~\cite{mirzadeh2024gsm} & 5,000 & \href{https://huggingface.co/datasets/apple/GSM-Symbolic}{apple/GSM-Symbolic} \\
        & GSM-Plus~\cite{li2024gsm} & 2,400 & \href{https://huggingface.co/datasets/qintongli/GSM-Plus}{qintongli/GSM-Plus} \\
        & OlympiadBench~\cite{he2024olympiadbench} & 674 & \href{https://huggingface.co/datasets/Hothan/OlympiadBench}{Hothan/OlympiadBench} \\
        & Omni-MATH~\cite{gao2024omni} & 4,428 & \href{https://huggingface.co/datasets/KbsdJames/Omni-MATH}{KbsdJames/Omni-MATH} \\
        & AMC2023 & Competition Math & \href{https://huggingface.co/datasets/math-ai/amc23}{math-ai/amc23} \\
        \midrule
        \textbf{Evaluation} & MMLU-Pro~\cite{wang2024mmlu} & 12,000 & \href{https://huggingface.co/datasets/TIGER-Lab/MMLU-Pro}{TIGER-Lab/MMLU-Pro} \\
        (General) & GPQA Diamond~\cite{rein2024gpqa} & 198 & \href{https://huggingface.co/datasets/Idavidrein/gpqa}{Idavidrein/gpqa} \\
        \bottomrule
    \end{tabular}
\end{table*}

\subsection{Training Infrastructure}
All experiments are conducted on a cluster of 8 NVIDIA H100 GPUs. To balance the computational load between policy evolution and environment feedback:
\begin{itemize}
    \item \textbf{Actor/Refiner}: Hosted on GPUs 0--3 for active policy gradient updates.
    \item \textbf{Evaluator/Solvers/Leak Judge}: Hosted on GPUs 4--7 to perform parallel inference for reward computation.
\end{itemize}
The training pipeline is implemented using PyTorch and the verl framework. We employ the GRPO algorithm for policy updates and utilize vLLM for accelerated offline reward inference.

\subsection{Hyperparameter Settings}
The detailed hyperparameters for the RL stage, ASH mechanism, and Leak Penalty are summarized in Table~\ref{tab:hyperparams}.

\begin{table}[t]
    \centering
    \caption{Hyperparameters for ReQueR Training.}
    \label{tab:hyperparams}
    \setlength{\tabcolsep}{1pt} 
    \small
    \begin{tabular}{l l r}
        \toprule
        \textbf{Category} & \textbf{Hyperparameter} & \textbf{Value} \\
        \midrule
        \textbf{RL (GRPO)} & Learning Rate & $1 \times 10^{-6}$ \\
        & Global Batch Size & 64 \\
        & PPO Mini-batch Size & 32 \\
        & Micro-batch Size per GPU & 8 \\
        & KL Coefficient ($\beta$) & 0.05 \\
        & Entropy Coefficient & 0 \\
        & Rollout Variants ($G$) & 8 \\
        & Total Training Epochs & 6 \\
        \midrule
        \textbf{ASH Mechanism} & Solver1 & Qwen3-0.6B \\
        & Solver2 & Qwen3-1.7B \\
        & Solver3 & Qwen3-4B \\
        & Initial Solver Label & $n_{\text{Solver}} // 2$ \\
        \midrule
        \textbf{Leak Penalty} & Perplexity Threshold ($\tau_{\text{leak}}$) & 5.0 \\
        & Penalty Coefficient ($\lambda$) & 1.0 \\
        & Score Deduction & -1.0 \\
        \bottomrule
    \end{tabular}
\end{table}

\subsection{Prompt Templates}
To ensure the reproducibility of our reasoning elicitation process, we detail the specific prompts used for both the Refiner and the downstream Solvers. Table~\ref{tab:refiner_prompt} illustrates the Refiner's meta-policy, while Table~\ref{tab:solver_prompt} defines the standardized evaluation environment for the Solvers.

\begin{table*}[t]
    \centering
    \caption{Refiner Prompt Template (SFT and RL Phases).}
    \label{tab:refiner_prompt}
    \small
    \renewcommand{\arraystretch}{1.2}
    \begin{tabularx}{\textwidth}{X}
        \toprule
        \cellcolor{gray!10}\textbf{System Prompt} \\
        \midrule
        \textbf{\# Role} \\
        You are an expert \textbf{Reasoning Query Refiner}. Your core task is \textbf{not} to answer questions directly, but to act as a ``cognitive frontend'' between a human user and a downstream AI Solver. You must rewrite the user's Raw Query into a Refined Query that is semantically equivalent but formulated in a way that maximizes the Solver's success rate. \\
        \textbf{\# Objective} \\
        Translate natural language ambiguity into machine-solvable logic. Disambiguate intent, make implicit constraints explicit, and structure the query to trigger the Solver's latent reasoning capabilities. \\
        \textbf{\# Input/Output Format} \\
        Input: User's raw natural language query. \\
        Output: First, analyze how to refine the problem step by step in \texttt{<think>} tags, then provide the refined query in \texttt{<rephrase>} tags. \\

        Format: \\
        \texttt{<think>(step by step thinking on how to rephrase questions)</think><rephrase>(rephrased version of the problem)</rephrase>} \\
        \midrule
        \cellcolor{gray!10}\textbf{User Prompt Template} \\
        \midrule
        Original Problem: \\
        \texttt{\{original\_question\}} \\
        Rephrase Problem: \\
        \bottomrule
    \end{tabularx}
\end{table*}

\begin{table*}[t]
    \centering
    \caption{Solver Prompt Template and Standardized Input Format.}
    \label{tab:solver_prompt}
    \small
    \renewcommand{\arraystretch}{1.2}
    \begin{tabularx}{\textwidth}{X}
        \toprule
        \cellcolor{gray!10}\textbf{System Prompt} \\
        \midrule
        You are a math assistant. Please solve the following problem step by step and put your final answer in \textbackslash{}boxed\{\} format. \\
        \midrule
        \cellcolor{gray!10}\textbf{User Input} \\
        \midrule
        Question: \\
        \texttt{\{problem\}} \\
        Answer: \\
        \bottomrule
    \end{tabularx}
\end{table*}

\section{Additional Experimental Analysis}
\label{appendix:additional}

To further characterize the behavior of ReQueR, we provide a set of supplementary analyses covering weak-to-strong transfer, cross-family curriculum, the role of supervised warm-up, the robustness of the leakage penalty, and failure-mode diagnostics.

\subsection{Weak-to-Strong Transfer Matrix}
\label{appendix:weak_to_strong}

While the main results (Table~\ref{tab:main_results1}) already implicitly demonstrate weak-to-strong transfer—the training pool caps at Qwen3-4B yet the Refiner consistently improves Solvers up to Qwen2.5-72B—we provide a controlled single-Solver analysis here. We train three variants of the Refiner, each using exactly one Solver during RL (Qwen3-0.6B, Qwen3-1.7B, or Qwen3-4B), and evaluate every variant on three unseen stronger Solvers across three benchmarks.

\begin{table*}[!t]
    \centering
    \caption{Weak-to-strong transfer across single-Solver Refiners and ASH. Every variant, including one trained solely on a 0.6B Solver, improves Solvers up to $120\times$ larger.}
    \label{tab:weak_to_strong}
    \small
    \setlength{\tabcolsep}{2.5pt}
    \begin{tabular}{llccccc}
        \toprule
        \textbf{Task} & \textbf{Test Solver} & \textbf{CoT} & \textbf{Only-0.6B} & \textbf{Only-1.7B} & \textbf{Only-4B} & \textbf{ASH} \\
        \midrule
        \multirow{3}{*}{MATH-500}
            & Qwen3-8B        & 83.40 & 83.85 & 83.65 & 83.25 & \textbf{85.00} \\
            & Llama-3.1-70B   & 64.60 & 71.25 & 70.75 & 70.85 & \textbf{72.00} \\
            & Qwen2.5-72B     & 76.20 & 79.85 & 78.40 & 80.10 & \textbf{83.00} \\
        \midrule
        \multirow{3}{*}{GSM-Symb.}
            & Qwen3-8B        & 86.96 & 87.18 & 87.38 & 87.64 & \textbf{87.08} \\
            & Llama-3.1-70B   & 85.76 & 86.54 & 86.70 & 86.96 & \textbf{87.98} \\
            & Qwen2.5-72B     & 82.26 & 87.72 & 87.98 & 87.38 & \textbf{88.12} \\
        \midrule
        \multirow{3}{*}{Olym.Bench}
            & Qwen3-8B        & 50.00 & 51.48 & 51.11 & 51.59 & \textbf{51.45} \\
            & Llama-3.1-70B   & 31.31 & 35.76 & 35.50 & 35.13 & \textbf{36.05} \\
            & Qwen2.5-72B     & 43.62 & 46.28 & 45.48 & 46.11 & \textbf{46.44} \\
        \midrule
        \multicolumn{2}{l}{\textit{Avg.}} & 67.12 & 69.99 & 69.66 & 69.89 & \textbf{70.79} \\
        \bottomrule
    \end{tabular}
\end{table*}

Several observations emerge. First, even the Only-0.6B Refiner—trained exclusively against a 0.6B Solver—improves Qwen2.5-72B by roughly $+3.9\%$ on average, spanning a $120\times$ scale gap. This indicates that the refinement policy captures universal reasoning triggers rather than model-specific shortcuts. Second, every single-Solver variant outperforms the CoT baseline on all unseen Solvers, confirming that the refinement generalizes upward regardless of training-Solver strength. Third, the multi-Solver ASH curriculum consistently achieves the best overall accuracy, validating its role in discovering more robust, architecture-agnostic triggers.

\subsection{Cross-Family Adaptive Solver Hierarchy}
\label{appendix:ash_cf}

Our main ASH uses Qwen3-0.6B/1.7B/4B to obtain a strictly monotonic capability ordering, which simplifies the curriculum update rule in Equation~\ref{eq:ash}. To verify that this design does not implicitly encode family-specific biases, we train a cross-family variant (ASH-CF) that replaces Qwen3-0.6B with Llama-3.2-3B-Instruct and keeps Qwen3-1.7B and Qwen3-4B unchanged.

\begin{table*}[!t]
    \centering
    \caption{Cross-family ASH (ASH-CF) compared with the single-family ASH across three benchmarks and six Solvers.}
    \label{tab:ash_cf}
    \small
    \setlength{\tabcolsep}{3pt}
    \begin{tabular}{llcccccc c}
        \toprule
        \textbf{Task} & \textbf{Method} & \makecell[c]{\textbf{Qwen3-}\\\textbf{0.6B}} & \makecell[c]{\textbf{Qwen3-}\\\textbf{1.7B}} & \makecell[c]{\textbf{Qwen3-}\\\textbf{8B}} & \makecell[c]{\textbf{Llama-3.2-}\\\textbf{3B-Inst.}} & \makecell[c]{\textbf{Llama-3.1-}\\\textbf{70B-Inst.}} & \makecell[c]{\textbf{Qwen2.5-}\\\textbf{72B-Inst.}} & \textbf{Avg.} \\
        \midrule
        \multirow{2}{*}{GSM-Symb.} & ASH    & 66.14 & 78.64 & 87.08 & 69.38 & 87.98 & 88.12 & 79.56 \\
                                   & ASH-CF & 66.06 & 79.10 & 87.74 & 69.84 & 87.86 & 88.38 & 79.83 \\
        \midrule
        \multirow{2}{*}{GSM-Plus}  & ASH    & 61.57 & 76.71 & 88.05 & 65.76 & 83.38 & 85.43 & 76.82 \\
                                   & ASH-CF & 64.00 & 77.38 & 86.52 & 64.81 & 84.62 & 85.86 & 77.20 \\
        \midrule
        \multirow{2}{*}{Olym.Bench}& ASH    & 21.66 & 39.02 & 51.45 & 19.88 & 36.05 & 46.44 & 35.75 \\
                                   & ASH-CF & 20.88 & 38.98 & 51.34 & 20.36 & 36.50 & 47.26 & 35.89 \\
        \midrule
        \multicolumn{2}{l}{Avg. (ASH)}      & 49.79 & 64.79 & 75.53 & 51.67 & 69.14 & 73.33 & 64.04 \\
        \multicolumn{2}{l}{Avg. (ASH-CF)}   & 50.31 & 65.15 & 75.20 & 51.67 & 69.66 & 73.83 & 64.31 \\
        \bottomrule
    \end{tabular}
\end{table*}

The two configurations perform comparably (64.04 vs.\ 64.31 on average), with ASH-CF offering marginal gains on cross-family Solvers such as Llama-3.1-70B and Qwen2.5-72B. This suggests that single-family ASH already captures architecture-agnostic reasoning triggers, and cross-family diversity primarily yields incremental benefits rather than correcting a systematic bias. We therefore keep the single-family design in the main experiments for training simplicity.

\subsection{Role of SFT: From-Scratch RL and Data Composition}
\label{appendix:sft_role}

The cold-start SFT dataset is composed as follows: 4{,}096 samples from OpenHermes-2.5, 1{,}024 samples from GSM8K (train), and 1{,}024 samples from MATH (train), yielding 6{,}144 samples in total. OpenHermes provides general-domain refinement patterns to prevent the Refiner from overfitting to mathematical phrasing, while the math subsets align the Refiner with the downstream RL training distribution.

Importantly, SFT in ReQueR is not the main source of reasoning capability; it mainly serves as format alignment, teaching the \texttt{<think>...</think><rephrase>...</rephrase>} output structure. To verify this, we train a Refiner using RL only, without any SFT initialization, and evaluate it across six benchmarks (MATH-500, GSM-Plus, GSM-Symbolic, OlympiadBench, AMC23, GPQA-Diamond) and three unseen stronger Solvers (Qwen3-8B, Llama-3.1-70B, Qwen2.5-72B).

\begin{table}[!t]
    \centering
    \caption{Effect of the SFT warm-up stage. From-scratch RL alone already achieves substantial gains; SFT accelerates convergence but is not essential.}
    \label{tab:from_scratch_rl}
    \small
    \begin{tabular}{lc}
        \toprule
        \textbf{Method} & \textbf{Avg. Accuracy} \\
        \midrule
        CoT (baseline)                 & 65.01 \\
        From-Scratch RL (no SFT)       & 67.72 \,(+2.71) \\
        Full ReQueR (SFT + RL)         & \textbf{68.01} \,(+3.00) \\
        \bottomrule
    \end{tabular}
\end{table}

From-scratch RL already closes most of the gap to full ReQueR, indicating that the reasoning-eliciting behavior is primarily acquired during the RL stage, and that ReQueR does not fundamentally depend on high-quality distillation data such as that generated by DeepSeek-R1.

\subsection{Leakage Penalty: Threshold Selection and Robustness}
\label{appendix:leak_robust}

\paragraph{Principled threshold selection.}
We construct a controlled leakage-detection benchmark by pairing positive samples (legitimate refinements) with negative samples (refinements with injected answer leakage) on MATH, and evaluate precision, recall, and F1 across candidate thresholds.

\begin{table*}[!t]
    \centering
    \caption{Detection performance of the perplexity-drop ratio across thresholds. $\tau=5.0$ achieves the peak F1.}
    \label{tab:leak_threshold_f1}
    \small
    \setlength{\tabcolsep}{6pt}
    \begin{tabular}{lccccccc}
        \toprule
        \textbf{Metric} & 1.5 & 2.0 & 3.0 & 4.0 & 5.0 & 7.0 & 10.0 \\
        \midrule
        Precision & 0.828 & 0.885 & 0.947 & 0.967 & \textbf{0.978} & 0.979 & 0.981 \\
        Recall    & 0.963 & 0.927 & 0.891 & 0.873 & 0.868 & 0.697 & 0.502 \\
        F1        & 0.890 & 0.906 & 0.918 & 0.918 & \textbf{0.920} & 0.814 & 0.664 \\
        \bottomrule
    \end{tabular}
\end{table*}

We further validate this choice by training separate Refiners with $\tau \in \{2.0, 3.0, 5.0, 7.0\}$ and evaluating across five benchmarks and five Solvers.

\begin{table*}[!t]
    \centering
    \caption{Downstream sensitivity to the leakage threshold $\tau$. Performance is stable across $\tau \in [2.0, 7.0]$, with $\tau=5.0$ giving the best average.}
    \label{tab:leak_threshold_downstream}
    \small
    \setlength{\tabcolsep}{6pt}
    \begin{tabular}{cccccc c}
        \toprule
        $\tau$ & \textbf{Qwen3-1.7B} & \textbf{Qwen3-4B} & \textbf{Qwen3-8B} & \textbf{Llama-3.1-70B} & \textbf{Qwen2.5-72B} & \textbf{Avg.} \\
        \midrule
        2.0 & 60.65 & 68.91 & 70.61 & 64.99 & 70.86 & 67.20 \\
        3.0 & 60.03 & 68.82 & 69.32 & 64.51 & 70.82 & 66.70 \\
        5.0 & 60.96 & 68.83 & \textbf{71.41} & \textbf{65.55} & \textbf{71.00} & \textbf{67.55} \\
        7.0 & 61.20 & 68.35 & 70.15 & 65.54 & 69.77 & 67.00 \\
        \bottomrule
    \end{tabular}
\end{table*}

The results remain stable in the $\tau \in [2.0, 7.0]$ range, confirming robustness to the specific threshold and supporting the use of $\tau=5.0$ in the main experiments.

\paragraph{Detecting paraphrased leakage.}
A key advantage of the perplexity-drop ratio over rule-based string matching is the ability to detect semantically paraphrased leakage. Consider the problem ``Let $p(x)=x^3-3x+1$. How many real roots does $p(p(x))=0$ have?'', whose ground-truth answer is 7. For three paraphrased leakage variants, the ratio is consistently well above $\tau=5.0$:

\begin{table*}[!t]
    \centering
    \caption{Paraphrased leakage is reliably detected through perplexity-drop, regardless of surface form.}
    \label{tab:paraphrased_leakage}
    \small
    \setlength{\tabcolsep}{6pt}
    \begin{tabular}{p{8cm}cccc}
        \toprule
        \textbf{Paraphrased Leakage} & $\text{PPL}(y^*|x)$ & $\text{PPL}(y^*|x')$ & \textbf{Ratio} & \textbf{Detected} \\
        \midrule
        ``The answer equals $2^3-1$.'' & 35.81 & 2.05 & 17.5 & \checkmark \\
        ``It can be shown that the answer is seven.'' & 35.81 & 1.10 & 32.6 & \checkmark \\
        ``Since $p$ has three real roots...giving $3+3+1=7$ total.'' & 35.81 & 1.83 & 19.6 & \checkmark \\
        \bottomrule
    \end{tabular}
\end{table*}

\paragraph{Naturally predictable answers.}
For well-posed queries with inherently predictable answers, the penalty does not produce false positives. For example, with ``What is $1+1$?'' rephrased as ``Compute the sum of 1 and 1.'', we obtain $\text{PPL}(y^*|x) = 1.26$ and $\text{PPL}(y^*|x') = 1.08$, giving a ratio of $1.17 \ll \tau = 5.0$. Because the penalty measures the \emph{relative} change in answer predictability rather than the absolute level, already-unambiguous queries cannot be incorrectly flagged as leakage.

\subsection{Failure Mode Diagnosis}
\label{appendix:failure_modes}

To understand when ReQueR fails, we manually categorize all failure cases (reward $=0$) on the MATH-500 validation set.

\begin{table*}[!t]
    \centering
    \caption{Failure-mode breakdown on MATH-500. The dominant mode reflects inherent Solver capability limits rather than Refiner errors.}
    \label{tab:failure_modes}
    \small
    \setlength{\tabcolsep}{6pt}
    \begin{tabular}{p{6cm}cp{8cm}}
        \toprule
        \textbf{Failure Category} & \textbf{Proportion} & \textbf{Description} \\
        \midrule
        Valid rephrase, Solver still fails & 85.2\% & Refinement is structurally sound; problem exceeds the Solver's capability. \\
        Diagram-dependent problems (\texttt{[asy]}) & 10.9\% & Visual information encoded in Asymptote blocks challenges the text-only Refiner. \\
        Low-quality rephrase & 3.9\% & Overly brief or verbose output, near-verbatim repetition, or formatting issues. \\
        \bottomrule
    \end{tabular}
\end{table*}

The majority of failures (85.2\%) are not caused by refinement errors: the refined query is structurally sound, but the problem's difficulty exceeds the Solver's latent reasoning ability. This directly aligns with our Limitations section, which states that ReQueR can only activate latent potential rather than create reasoning capabilities absent from the Solver. The remaining failures are bounded and predictable: diagram-dependent problems whose reasoning hinges on visual elements, and a small fraction of low-quality outputs.

\subsection{Self-Improving Refiner via Emergent Strategies}
\label{appendix:self_improve}

As an exploratory extension, we test whether the emergent refinement strategies discovered by the Refiner itself (Tables~\ref{tab:prompt_strategies} and~\ref{tab:prompt_strategies_full}) can be fed back into its own prompt to form a self-improving loop. We inject the seven emergent strategies as explicit heuristic guidance into the Refiner's system prompt and retrain; evaluation uses Qwen3-0.6B as the Solver.

\begin{table}[!t]
    \centering
    \caption{Recursive refinement: using the Refiner's own emergent strategies as prompt guidance yields further gains.}
    \label{tab:self_improve}
    \small
    \setlength{\tabcolsep}{4pt}
    \begin{tabular}{lccc}
        \toprule
        \textbf{Benchmark} & \textbf{Simple Prompt} & \textbf{Strategy Prompt} & $\Delta$ \\
        \midrule
        AMC23        & 23.75 & 25.62 & +1.87 \\
        Olym.Bench   & 21.66 & 24.81 & +3.15 \\
        GSM-Plus     & 61.57 & 64.76 & +3.19 \\
        GSM-Symb.    & 66.14 & 66.54 & +0.40 \\
        MATH-500     & 54.40 & 55.85 & +1.45 \\
        \bottomrule
    \end{tabular}
\end{table}

The Strategy Prompt improves performance across all five benchmarks, suggesting that the Refiner's discoveries can be recycled to bootstrap an even stronger Refiner. We view this as a promising direction for future work on recursive self-improvement of inference-time meta-policies.

\section{Qualitative Analysis: Refinement Archetypes and Case Studies}
\label{app:qualitative}

\subsection{Comprehensive Taxonomy of Refinement Strategies}
To investigate the operational mechanisms of the Refiner's meta-policy, we perform a qualitative decomposition of the generated queries across diverse reasoning tasks. Table~\ref{tab:prompt_strategies_full} provides a comprehensive taxonomy of seven emergent strategies that facilitate the transition from raw human intuition to machine-executable logic. These strategies, ranging from \textit{Symbolic Parameterization} to \textit{Chain-of-Thought Structuring}, do not merely rephrase the input text but perform a \textit{Structural Transduction} of the problem space. By rectifying specific logical vulnerabilities and making implicit constraints explicit, these archetypes ensure that the refined query is projected onto a logic-dense manifold, effectively eliciting the latent reasoning capabilities of downstream Solvers regardless of their architecture or scale.
\definecolor{MorandiBlue}{RGB}{163, 184, 204}
\definecolor{DeepBlue}{RGB}{70, 90, 110}

\definecolor{MorandiGreen}{RGB}{168, 181, 158}
\definecolor{DeepGreen}{RGB}{80, 95, 70}

\definecolor{MorandiPink}{RGB}{205, 175, 175}
\definecolor{DeepPink}{RGB}{120, 80, 80}

\definecolor{MorandiTan}{RGB}{214, 196, 161}
\definecolor{DeepTan}{RGB}{110, 90, 50}

\definecolor{MorandiPurple}{RGB}{180, 165, 190}
\definecolor{DeepPurple}{RGB}{90, 75, 100}

\definecolor{MorandiOrange}{RGB}{220, 190, 170}
\definecolor{DeepOrange}{RGB}{130, 90, 70}

\definecolor{MorandiTeal}{RGB}{160, 190, 190}
\definecolor{DeepTeal}{RGB}{60, 100, 100}

\begin{table*}[t]
\centering
\caption{Comprehensive taxonomy of emergent strategies for converting human intuition into machine-executable logic.}
\vspace{-8pt}
\label{tab:prompt_strategies_full}

% --- Strategy 1: Symbolic Parameterization ---
\begin{strategybox}{Strategy 1: Symbolic Parameterization}{MorandiBlue}
    \scriptsize \textbf{Description:} Transforms narrative word problems into formal symbolic models comprising variables and equations.
    \vspace{2pt} \hrule \vspace{5pt}
    \begin{tabularx}{\textwidth}{l|X|X}
        & \textbf{Raw Query} & \textbf{Refined Query} \\ \hline
        \textbf{Ex.} & Ten treeks weigh as much as three squigs and one goolee. Two treeks and one goolee are equal in weight to one squig. The combined weight of how many treeks equals the weight of one squig? & 
        \textbf{\textcolor{DeepBlue}{Let $T$ be the weight of one treek, $S$ be the weight of one squig, and $G$ be the weight of one goolee.}} We are given: \textbf{\textcolor{DeepBlue}{1. $10T = 3S + G$, 2. $2T + G = S$.}} We need to \textbf{\textcolor{DeepBlue}{find the number $n$ such that $nT = S$}}. Determine $n$ by solving the system of equations. \\
    \end{tabularx}
\end{strategybox}

% --- Strategy 2: Contextual De-noising ---
\begin{strategybox}{Strategy 2: Contextual De-noising}{MorandiGreen}
    \scriptsize \textbf{Description:} Filters out factual but computationally irrelevant narrative noise from the problem description.
    \vspace{2pt} \hrule \vspace{5pt}
    \begin{tabularx}{\textwidth}{l|X|X}
        & \textbf{Raw Query} & \textbf{Refined Query} \\ \hline
        \textbf{Ex.} & \stcolor{DeepGreen}{Daniel works at an electronics store, and he claims that} the popularity of a television is inversely proportional to its cost. If 15 customers buy a television that costs \$1500, how many customers would buy a television that costs \$2500? & 
        \textbf{\textcolor{DeepGreen}{Given that}} the number of customers buying a television is inversely proportional to its cost, and that 15 customers buy a television costing \$1500, how many customers would buy a television costing \$2500? \\
    \end{tabularx}
\end{strategybox}

% --- Strategy 3: Linguistic Disambiguation ---
\begin{strategybox}{Strategy 3: Linguistic Disambiguation}{MorandiPink}
    \scriptsize \textbf{Description:} Resolves semantic ambiguity by replacing vague natural language with formal operators or explicit sets.
    \vspace{2pt} \hrule \vspace{5pt}
    \begin{tabularx}{\textwidth}{l|X|X}
        & \textbf{Raw Query} & \textbf{Refined Query} \\ \hline
        \textbf{Ex.} & What is the \wavecolor{DeepPink}{sum of all of the multiples of 3 between 100 and 200}? & 
        Calculate the sum of the \textbf{\textcolor{DeepPink}{arithmetic series}} starting at \textbf{\textcolor{DeepPink}{102}}, ending at \textbf{\textcolor{DeepPink}{198}}, with a \textbf{\textcolor{DeepPink}{common difference of 3}}. \\
    \end{tabularx}
\end{strategybox}

% --- Strategy 4: Conceptual Reduction ---
\begin{strategybox}{Strategy 4: Conceptual Reduction}{MorandiTan}
    \scriptsize \textbf{Description:} Decomposes high-level math concepts into primitive algebraic operations and decision constraints.
    \vspace{2pt} \hrule \vspace{5pt}
    \begin{tabularx}{\textwidth}{l|X|X}
        & \textbf{Raw Query} & \textbf{Refined Query} \\ \hline
        \textbf{Ex.} & How many \wavecolor{DeepTan}{vertical asymptotes} does the graph of $y = \frac{2}{x^2+x-6}$ have? & 
        Find the number of \textbf{\textcolor{DeepTan}{distinct real numbers $x$}} for which the \textbf{\textcolor{DeepTan}{denominator}} of the function $y = \frac{2}{x^2+x-6}$ \textbf{\textcolor{DeepTan}{is zero}}. \\
    \end{tabularx}
\end{strategybox}

% --- Strategy 5: Procedural Hint Injection ---
\begin{strategybox}{Strategy 5: Procedural Hint Injection}{MorandiPurple}
    \scriptsize \textbf{Description:} Augments the query with necessary domain formulas or constants to minimize Solver-side retrieval failures.
    \vspace{2pt} \hrule \vspace{5pt}
    \begin{tabularx}{\textwidth}{l|X|X}
        & \textbf{Raw Query} & \textbf{Refined Query} \\ \hline
        \textbf{Ex.} & Evaluate $i^5 + i^{-25} + i^{45}$ & 
        Evaluate the expression $i^5 + i^{-25} + i^{45}$ by \textbf{\textcolor{DeepPurple}{simplifying each term using the cyclical pattern}} of the powers of the imaginary unit $i$, which repeats every four exponents: \textbf{\textcolor{DeepPurple}{$i^0 = 1, i^1 = i, i^2 = -1, i^3 = -i$}}. \\
    \end{tabularx}
\end{strategybox}

% --- Strategy 6: Explicit Grounding ---
\begin{strategybox}{Strategy 6: Explicit Grounding}{MorandiOrange}
    \scriptsize \textbf{Description:} Elevates implicit mathematical axioms (e.g., non-zero denominators) to mandatory directives.
    \vspace{2pt} \hrule \vspace{5pt}
    \begin{tabularx}{\textwidth}{l|X|X}
        & \textbf{Raw Query} & \textbf{Refined Query} \\ \hline
        \textbf{Ex.} & For what value of $x$ will $\frac{2x-1}{2x+2}$ and $\frac{x-3}{x-1}$ be equal? & 
        Solve the equation \textbf{\textcolor{DeepOrange}{$\frac{2x-1}{2x+2} = \frac{x-3}{x-1}$}} for $x$, \textbf{\textcolor{DeepOrange}{ensuring to exclude any values that make the denominators zero}}. Find the value of $x$ that satisfies this equation. \\
    \end{tabularx}
\end{strategybox}

% --- Strategy 7: Chain-of-Thought Structuring ---
\begin{strategybox}{Strategy 7: Chain-of-Thought Structuring}{MorandiTeal}
    \scriptsize \textbf{Description:} Enforces a sequential reasoning path via step-by-step instructions, converting latent reasoning into explicit execution.
    \vspace{2pt} \hrule \vspace{5pt}
    \begin{tabularx}{\textwidth}{l|X|X}
        & \textbf{Raw Query} & \textbf{Refined Query} \\ \hline
        \textbf{Ex.} & Given the functions $f(x) = \frac{x+5}{3}$ and $g(x) = \frac{1}{f^{-1}(x) + 1}$, find the value of $g(3)$. & 
        Given the function $f(x) = \frac{x+5}{3}$, \textbf{\textcolor{DeepTeal}{first determine its inverse function}} $f^{-1}(x)$. Then, \textbf{\textcolor{DeepTeal}{define a new function}} $g(x) = \frac{1}{f^{-1}(x) + 1}$. \textbf{\textcolor{DeepTeal}{Finally, compute the value of $g(3)$}}. \\
    \end{tabularx}
\end{strategybox}

\vspace{-5pt}
\end{table*}

\subsection{Generalization to Unseen Domains}
\begin{table*}[t]
\centering
\caption{Cross-Domain Generalization Case Studies in MMLU-Pro: ReQueR Transforming Human Queries into Structured Logic.}
\vspace{-8pt}
\label{tab:mmlu_cases}

% --- Case 1: Law (Strategies 2, 4) ---
\begin{strategybox}{Domain: Law (De-noising \& Conceptual Reduction)}{MorandiGreen}
    \scriptsize \textbf{Mechanism:} Filters personal anecdotes (editor's daughter) and maps the dispute to the constitutional intersection of Commercial Speech vs. Private Editorial Control.
    \vspace{2pt} \hrule \vspace{5pt}
    \begin{tabularx}{\textwidth}{l|X|X}
        & \textbf{Raw Query)} & \textbf{Refined Query} \\ \hline
        \textbf{Ex.} & (Narrative about a newspaper editor refusing a test-prep ad because his daughter failed the course and now works at the mall)... In an action to require the newspaper to publish, who will prevail? & 
        Analyze whether a private newspaper can be compelled to publish an advertisement under the \textbf{\textcolor{DeepGreen}{First Amendment commercial speech doctrine}}. Evaluate the \textbf{\textcolor{DeepGreen}{private actor's editorial discretion}} against the owner's claim for access, considering that no state action is involved. \\
    \end{tabularx}
\end{strategybox}

% --- Case 2: Biology (Strategies 3, 7) ---
\begin{strategybox}{Domain: Biology (Linguistic Disambiguation \& CoT Structuring)}{MorandiPink}
    \scriptsize \textbf{Mechanism:} Converts a general ``Explain'' prompt into a rigorous experimental design task with a clear methodological sequence.
    \vspace{2pt} \hrule \vspace{5pt}
    \begin{tabularx}{\textwidth}{l|X|X}
        & \textbf{Raw Query} & \textbf{Refined Query} \\ \hline
        \textbf{Ex.} & Using conditioning, how can one determine the sensitivity of a dog's hearing to sounds of different frequencies? & 
        \textbf{\textcolor{DeepPink}{Design an experimental procedure using classical conditioning}} to measure hearing sensitivity. \textbf{\textcolor{DeepPink}{First, describe the pairing}} of specific frequencies with a stimulus. \textbf{\textcolor{DeepPink}{Then, outline how to test sensitivity limits}} by varying frequency while holding other environmental variables constant. \\
    \end{tabularx}
\end{strategybox}

% --- Case 3: Business (Strategies 1, 7) ---
\begin{strategybox}{Domain: Business (Symbolic Parameterization \& CoT Structuring)}{MorandiBlue}
    \scriptsize \textbf{Mechanism:} Automatically parameterizes supply and demand functions, converting a market narrative into a solvable system of equations.
    \vspace{2pt} \hrule \vspace{5pt}
    \begin{tabularx}{\textwidth}{l|X|X}
        & \textbf{Raw Query} & \textbf{Refined Query} \\ \hline
        \textbf{Ex.} & Distributors are willing to buy $60-p$ barrels, while the Republic is willing to supply $p^2/10$. i) What is the equilibrium price? ii) How many barrels will be sold? & 
        \textbf{\textcolor{DeepBlue}{Let $D(p) = 60-p$ and $S(p) = p^2/10$}}. \textbf{\textcolor{DeepBlue}{First, find the equilibrium price $p$}} by setting the supply and demand functions equal ($S(p) = D(p)$). \textbf{\textcolor{DeepBlue}{Then, evaluate the equilibrium quantity}} by substituting $p$ back into either function. \\
    \end{tabularx}
\end{strategybox}

% --- Case 4: Engineering (Strategies 4, 5) ---
\begin{strategybox}{Domain: Engineering (Conceptual Reduction \& Hint Injection)}{MorandiPurple}
    \scriptsize \textbf{Mechanism:} Bridges the gap between a physical property description and a mathematical operator ($\frac{Z_L - Z_0}{Z_L + Z_0}$).
    \vspace{2pt} \hrule \vspace{5pt}
    \begin{tabularx}{\textwidth}{l|X|X}
        & \textbf{Raw Query} & \textbf{Refined Query} \\ \hline
        \textbf{Ex.} & Find the reflection coefficient for voltage waves at the load end of a 50 ohm transmission line terminated with a load impedance of $25 - j75$ ohms. & 
        Given characteristic impedance \textbf{\textcolor{DeepPurple}{$Z_0 = 50 \Omega$}} and load impedance \textbf{\textcolor{DeepPurple}{$Z_L = 25 - j75 \Omega$}}, compute the reflection coefficient $\Gamma$ \textbf{\textcolor{DeepPurple}{using the formula $\Gamma = (Z_L - Z_0) / (Z_L + Z_0)$}}. Express the result in polar form. \\
    \end{tabularx}
\end{strategybox}

% --- Case 5: Computer Science (Strategy 6) ---
\begin{strategybox}{Domain: Computer Science (Explicit Grounding)}{MorandiOrange}
    \scriptsize \textbf{Mechanism:} Identifies the specific logical failure mode (post-target iteration) to ground the debugging task in precise algorithmic behavior.
    \vspace{2pt} \hrule \vspace{5pt}
    \begin{tabularx}{\textwidth}{l|X|X}
        & \textbf{Raw Query} & \textbf{Refined Query} \\ \hline
        \textbf{Ex.} & Which of the following procedure calls can be used to demonstrate that the procedure does NOT Work as intended? (Followed by pseudocode) & 
        The procedure contains a logic flaw where \textbf{\textcolor{DeepOrange}{foundIndex is reset to 0 if a non-target name is encountered after the match}}. Identify a test case that \textbf{\textcolor{DeepOrange}{specifically triggers this failure}} by placing the target name in a non-terminal position in the list. \\
    \end{tabularx}
\end{strategybox}

% --- Case 6: Economics (Strategy 6) ---
\begin{strategybox}{Domain: Economics (Boundary Limitation)}{MorandiTan}
    \scriptsize \textbf{Mechanism:} Explicitly incorporates the ``ceteris paribus'' condition (keeping other factors constant) to satisfy formal economic definitions and logical boundaries.
    \vspace{2pt} \hrule \vspace{5pt}
    \begin{tabularx}{\textwidth}{l|X|X}
        & \textbf{Raw Query} & \textbf{Refined Query} \\ \hline
        \textbf{Ex.} & What is meant by the marginal product of a productive factor? & 
        Define the marginal product of a productive factor in economics, specifying that it refers to the additional output produced by an additional unit of that factor \textbf{\textcolor{DeepTan}{while keeping all other factors constant}}. \\
    \end{tabularx}
\end{strategybox}

% --- Case 7: Physics (Strategy 5) ---
\begin{strategybox}{Domain: Physics (Hint Injection)}{MorandiTeal}
    \scriptsize \textbf{Mechanism:} Proactively supplies missing physical constants (e.g., gravitational acceleration) when handling imperial units to eliminate Solver uncertainty.
    \vspace{2pt} \hrule \vspace{5pt}
    \begin{tabularx}{\textwidth}{l|X|X}
        & \textbf{Raw Query} & \textbf{Refined Query} \\ \hline
        \textbf{Ex.} & In what distance can a 3000-lb automobile be stopped from a speed of 30 mi/hr (44 ft/sec) if the coefficient of friction is 0.70? & 
        Given a car with a weight of 3000 lb and an initial speed of 44 ft/s, calculate the stopping distance. \textbf{\textcolor{DeepTeal}{Assume standard gravitational acceleration of $32 \text{ ft/s}^2$}} to resolve unit-based ambiguity. \\
    \end{tabularx}
\end{strategybox}

\vspace{-5pt}
\end{table*}

The effectiveness of the Refiner meta-policy is highly suggestive of its capacity to internalize a domain-invariant reasoning syntax. Rather than merely memorizing task-specific patterns, the Refiner appears to utilize mathematical datasets such as GSM8K and MATH as a rigorous training ground to master structural primitives that are foundational to logical discourse. This cross-domain adaptability is empirically illustrated in Table~\ref{tab:mmlu_cases}, which details representative case studies curated from the MMLU-Pro benchmark. These examples demonstrate the Refiner's ability to perform Structural Transduction across diverse fields including law, biology, and computer science. 

By decoupling the process of reasoning from the repository of domain knowledge, the Refiner functions as a versatile cognitive frontend that prioritizes formal clarity over semantic surface. As observed in the MMLU-Pro evaluations, the symbolic parameterization strategy originally developed for mathematical word problems is functionally parallel to the process of isolating editorial discretion in a legal dispute, as both operations require the Refiner to strip away semantic noise to reveal the underlying logical constraints. Consequently, ReQueR establishes a standardized interface that enables downstream Solvers to mitigate linguistic ambiguity and more effectively activate their latent logical capabilities across the entire MMLU-Pro spectrum. This zero-shot generalization indicates that the Refiner has moved beyond simple pattern matching toward a deeper, model-agnostic understanding of problem decomposition.

% \section{Extended Empirical Results}

% \subsection{Complete Benchmark Performance}
% % Full evaluation tables containing all sub-dataset scores that were summarized in the main text.

% \section{Analysis of ASH and Leakage Penalty}

% \subsection{Solver Evolution Trajectories}
% % Visualization of Solver assignment dynamics across different task difficulty levels throughout the training process.

% \subsection{PPL Threshold Sensitivity}
% % Ablation study on the sensitivity of the leakage penalty to different perplexity thresholds, highlighting the trade-off between strictness and policy flexibility.
% % 这里可以放一张表格，展现不同的case，同时制造一批leak数据集，展现不同方法准确度和时间开销
% % LLM-as-Judge 速度慢（具体数据），且不稳定
% % rule-based 太呆板

% % PPL leak ratio 容忍度的自动初始化

\section{Theoretical Framework}
\label{sec:theory}

The ReQueR framework establishes a formal paradigm for reasoning elicitation by decoupling the alignment process from the Solver’s internal parametric space. We justify this approach through the dual lenses of Amortized Complexity and Information-Theoretic Feedback Calibration.

\subsection{The O(1) Paradigm: Amortized Alignment via Input-Side Mapping}

Traditional alignment methodologies, such as Supervised Fine-Tuning or Reinforcement Learning, operate primarily within the Parametric Manifold $\Theta$. For a heterogeneous ensemble of $N$ Solvers $\mathcal{S} = \{s_1, \dots, s_N\}$, where each $s_i$ is parameterized by $\theta_i \in \Theta_i$, the total parametric alignment cost $\mathcal{C}_{\text{param}}$ scales linearly with $N$:
\begin{equation}
\mathcal{C}_{\text{param}} = \sum_{i=1}^N \int_{t=0}^T \|\nabla_{\theta_i} \mathcal{J}(\theta_i, \mathcal{D})\| dt \in O(N),
\end{equation}
where $\mathcal{J}$ is the alignment objective and $T$ is the optimization horizon. 

Existing Automated Prompt Optimization techniques mitigate this overhead by performing a Point-Estimate Optimization over the discrete token space $\mathcal{V}^*$. APO seeks a static wrapper string $p^*$ for a specific task distribution $\mathcal{D}_k$:
\begin{equation}
p^*_{APO} = \arg\max_{p \in \mathcal{V}^*} \mathbb{E}_{x \sim \mathcal{D}_k} [\mathcal{R}(s(p \oplus x))],
\end{equation}
where $\oplus$ denotes string concatenation. While parameter-efficient, this ``one-size-fits-all'' wrapper lacks per-sample granularity and is often tightly coupled to the internal biases of a specific Solver, limiting its zero-shot migration capability.

In contrast, ReQueR reformulates alignment as an Amortized Mapping problem $\Phi: \mathcal{X} \to \mathcal{X}'$, where $\mathcal{X}'$ is a reasoning-dense manifold. Let $\pi_\theta(x'|x)$ be the Refiner policy. We define the universal alignment objective as maximizing the expectation over both the task distribution $\mathcal{D}$ and the Solver manifold $\mathcal{S}$:
\begin{equation}
\begin{aligned}
\max_{\theta} \mathcal{J}(\theta) = & \mathbb{E}_{x \sim \mathcal{D}} \bigg[ \int_{s \in \mathcal{S}} \mathbb{E}_{x' \sim \pi_\theta(x'|x)} \\
& \left[ \mathcal{R}(s(x')) \right] d\mu(s) \bigg],
\end{aligned}
\end{equation}
where $\mu(s)$ is a measure over the Solver space. Since $\theta$ is optimized once and remains invariant across all $s \in \mathcal{S}$, the complexity relative to the number of Solvers becomes $\mathcal{C}_{\text{ReQueR}} \in O(1)$. The Refiner functions as an Invariant Canonical Projection~\cite{jin2026invariant,ma2026gor}, extracting structural logical primitives that remain cross-architecturally effective.

\subsection{ASH: Optimization Stability via Gradient Signal Recovery}

The Adaptive Solver Hierarchy serves as a strategic framework for Gradient Signal Recovery. Under the GRPO framework, the policy gradient $\nabla_\theta \mathcal{J}(\theta)$ is driven by the group-relative advantage $A_i$:
\begin{equation}
\nabla_\theta \mathcal{J}(\theta) \approx \frac{1}{G} \sum_{i=1}^G \nabla_\theta \log \pi_\theta(x_i'|x) \cdot \frac{R_i - \bar{R}}{\sigma_R + \epsilon},
\end{equation}
where $\bar{R} = \frac{1}{G}\sum R_i$ and $\sigma_R$ is the group standard deviation.

\paragraph{Reward Entropy Collapse and SNR} 
For a binary reward $\mathcal{R} \in \{0, 1\}$, the informativeness of the update depends on the Signal-to-Noise Ratio (SNR) of the advantage signal:
\begin{equation}
\text{SNR} = \frac{\text{Var}(\mathbb{E}[R | \pi_\theta])}{\mathbb{E}[\text{Var}(R | \pi_\theta)] + \sigma_{\text{noise}}^2},
\end{equation}
When the Solver $s_k$ is poorly matched to the task difficulty (e.g., an ``Expert'' where $P \to 1$ or a ``Novice'' where $P \to 0$), the reward distribution suffers from Entropy Collapse~\cite{hao2025rethinking}, i.e., $\mathbb{H}(\mathcal{R}) \to 0$. Consequently, $\sigma_R \to 0$, leading to a vanishing SNR and gradient stagnation.

\paragraph{Optimal Calibration via ZPD Constraints} 
ASH operationalizes the Zone of Proximal Development by dynamically selecting a Solver $s_{k^*}$ that acts as a Maximum Information Elicitor. We define the optimal Solver index at time $t$ as:
\begin{equation}
\begin{aligned}
k^*(t) = & \arg\min_{k} \Big\| \mathbb{E}_{x \sim \mathcal{D}} \\
& \left[ P(y = y^* | x, s_k, \pi_\theta) \right] - \tau \Big\|_2,
\end{aligned}
\end{equation}
where $\tau \in (0, 1)$ is the target difficulty threshold (typically $\tau \approx 0.5$ to maximize the variance $\tau(1-\tau)$). By maintaining the Refiner at its ZPD edge, ASH ensures Continuous Gradient Pressure, forcing the synthesis of ``Hardened Triggers'' that possess high logical density and are cross-architecturally effective.

\section{Promising Application Scenarios}
\label{appendix:scenarios}

Beyond the benchmarks presented in this study, the $O(1)$ alignment paradigm of ReQueR offers significant potential as a Cognitive Frontend that bridges the gap between raw human intent and the specialized execution logic of heterogeneous Solvers. By decoupling \textit{reasoning elicitation}~\cite{dong2026neureasoner} from \textit{knowledge retrieval}, ReQueR enables a scalable Edge-Cloud Collaborative Inference architecture where a centralized, high-capacity Refiner empowers a diverse fleet of edge-deployed models without individual fine-tuning.

\paragraph{Autonomous Code Agents as Requirement Architects} 
In the domain of complex software engineering, ReQueR transcends simple query rephrasing by functioning as a \textit{Requirement Architect}. By identifying implicit logical constraints and structural dependencies within ambiguous user specifications, the Refiner projects raw natural language into a ``logic-dense'' manifold. This process significantly minimizes the search space for downstream code-generation Solvers and mitigates the hallucination of non-existent API parameters~\cite{jiang2026foe}, allowing even small-scale code models to execute high-level architectural designs with precision.

\paragraph{Agentic Workflow Orchestration} 
For multi-agent systems requiring sophisticated tool-use and multi-step reasoning~\cite{lin2025se}, ReQueR serves as a universal task-decomposer. Instead of the prohibitive cost of fine-tuning every specialized agent for instruction following, the Refiner reformulates global objectives into a sequence of atomic, model-invariant sub-tasks. This orchestration ensures that heterogeneous agents can execute complex workflows with high fidelity, effectively lowering the barrier for deploying advanced agentic intelligence in resource-constrained or distributed environments.

\paragraph{Privacy-Preserving Edge Reasoning} 
ReQueR further facilitates a hybrid inference paradigm where data sovereignty and reasoning capacity are simultaneously optimized. A cloud-based Refiner can generate ``reasoning-enhanced'' meta-logical templates that are strictly independent of sensitive user content. These templates are subsequently combined with private data locally to guide a Small Language Model (SLM). Such a decoupling allows edge devices to inherit the reasoning depth of giant models~\cite{li2024towards} while ensuring that private information never leaves the local environment, defining a new frontier for secure, decentralized AI.

\section{Use of Large Language Models}
\label{appendix:use_of_llms}

We utilized large language models in the following aspects of this research:

\begin{itemize}
    \item \textbf{Data Synthesis:} DeepSeek-R1 was used to generate 6,144 high-quality instruction-following samples for the cold-start Supervised Fine-Tuning phase of the Refiner model.
    \item \textbf{Manuscript Preparation:} Gemini was employed for grammar and spell checking to improve the clarity and readability of the manuscript.
\end{itemize}

\end{document}